\documentclass{article}

\pdfoutput=1

\usepackage{arxiv}
\usepackage[utf8]{inputenc}
\usepackage[T1]{fontenc}   
\usepackage{hyperref}
\usepackage{url}
\usepackage{booktabs}
\usepackage{amsfonts} 
\usepackage{nicefrac}
\usepackage{microtype}
\usepackage{lipsum}
\usepackage{graphicx}
\usepackage{natbib}
\usepackage{doi}
\usepackage{siunitx}
\usepackage{amsmath}
\usepackage{cleveref}
\usepackage{enumitem}
\usepackage{subcaption}
\usepackage{multirow}

\title{Data-Adaptive Dimensional Analysis 
for Accurate Interpolation and Extrapolation in Computer Experiments}

\date{December 14, 2023}

\author{G. Alexi~Rodr\'{i}guez-Arelis \\
	Department of Statistics \\
	University of British Columbia \\
	Vancouver, BC V6T 1Z4, Canada \\
	\texttt{alexrod@stat.ubc.ca} \\
	\And
	William J.~Welch \\
	Department of Statistics \\
	University of British Columbia \\
	Vancouver, BC V6T 1Z4, Canada  \\
	\texttt{will@stat.ubc.ca} \\
}

\renewcommand{\shorttitle}{Data-Adaptive Dimensional Analysis}
\newcommand{\jrange}[2]{j = #1,\ldots,\allowbreak #2}
\newcommand{\xVec}{\mathbf{x}}
\newcommand{\subs}[3]{#1_{#2},\ldots,\allowbreak #1_{#3}}

\begin{document}
\maketitle

\begin{abstract}
Dimensional analysis (DA) pays attention to fundamental physical dimensions such as length and mass when modelling scientific and engineering systems. It goes back at least a century to Buckingham's $\Pi$ theorem, which characterizes a scientifically meaningful model in terms of a limited number of dimensionless variables. The methodology has only been exploited relatively recently by statisticians for design and analysis of experiments, however, and computer experiments in particular. The basic idea is to build models in terms of new dimensionless quantities derived from the original input and output variables. A scientifically valid formulation has the potential for improved prediction accuracy in principle, but the implementation of DA is far from straightforward. There can be a combinatorial number of possible models satisfying the conditions of the theory. Empirical approaches for finding effective derived variables will be described, and improvements in prediction accuracy will be demonstrated. As DA's dimensionless  quantities for a statistical model typically compare the original variables rather than use their absolute magnitudes, DA is less dependent on the choice of experimental ranges in the training data. Hence, we are also able to illustrate sustained accuracy gains even when extrapolating substantially outside the training data.
\end{abstract}


\section{Introduction}
\label{sec:intro}

In a wide range of natural phenomena and engineering processes, physical experimentation is resource-intensive or even impossible, motivating the now widespread use of mathematical models implemented as computer codes. This complementary way of doing science has for decades spawned corresponding research in statistical methodologies for the careful design and analysis of computer experiments \citep[DACE,][]{currin1991,sacks1989}. When a complex computer code is expensive to evaluate, DACE replaces the code by a fast statistical model surrogate trained with limited code runs. The surrogate most commonly employed treats the unknown input-output function as a realization of a Gaussian stochastic process (GaSP), also known simply as a Gaussian process (GP).

While there are many possible objectives of such an experiment, e.g., optimization or calibration, prediction of the original computer code output by the statistical surrogate underlies tackling the scientific objective. Therefore, obtaining good accuracy at untried values of the input variables is a fundamental goal. Typically, input variables are in a ``raw'' form, as provided to the code, and the output is similarly as produced by the code or some summary. This article maintains the basic GaSP surrogate paradigm but proposes to improve prediction accuracy by input and output transformations guided by dimensional analysis (DA), which is related to physical units of measurement (and not the number of variables). 

DA is a theory going back more than a century to \cite{buckingham1914} and others but has received scant attention from statisticians. \cite{finney1977} compared modeling in physics, where dimensional consistency of equations is routinely taught, versus the almost universal lack of attention to these fundamental concepts by statisticians. The topic has appeared occasionally in more recent statistical literature, however. \cite{AlbNacAlb2013} and \cite{shen2014} demonstrated the potential advantages of DA for design and analysis of physical experiments, including novel designs in derived DA variables. \cite{shen2018} introduced a conjugate method for formulating and incorporating DA in statistical models. For computer experiments, the topic of the current article, \cite{shen2018_2} proposed designs with good coverage in the domain of derived DA variables, determined before experimentation. An important distinction between these works and the current article is that we choose the DA variables empirically.

The rest of this article is organized as follows. Section~\ref{sec:buckingham} outlines the main ideas of DA in Buckingham's $\Pi$ theorem, with emphasis on the implications for statistical prediction, especially extrapolation. The review of standard DACE methodology in Section~\ref{sec:modeling} concentrates on the requirements of the functional analysis of variance (FANOVA) that guides our approach to DA. Section~\ref{sec:implementation} outlines the key steps to execute DA \citep{AlbNacAlb2013,shen2014} with some commentary about our new choices for the implementation. The two case studies in Sections~\ref{sec:b_function} and~\ref{sec:solid_sphere} elaborate on these steps for two applications where there is much modeling latitude within the constraints of DA and demonstrate the substantial gains in prediction accuracy possible. Finally, Section~\ref{sect:discussion} makes some concluding remarks and suggests further opportunities for improvement.

\section{Buckingham's $\Pi$ Theorem}
\label{sec:buckingham}

Here we outline a key DA result---Buckingham's $\Pi$ theorem---paying special attention to the implications for statistical modeling and extrapolation.

DA considers the dimensions of variables, and it is useful to clarify the term ``dimension''. It is not the number of variables, the usual definition in statistical science, but rather a fundamental dimension of measurement appearing in a physically meaningful system. The fundamental dimensions are mass ($\si{M}$), length ($\si{L}$), time ($\si{T}$), temperature ($\Theta$), electric current ($\si{Q}$), amount of substance ($\si{N}$), and luminous intensity ($\si{I_v}$). For example, the Pythagorean relationship 
\begin{equation}
\label{eqn:pythag}
y = \sqrt{x_1^2 + x_2^2}
\end{equation} 
for the length of the hypotenuse of a right-angle triangle as a function of the lengths $x_1$ and $x_2$ of the other two sides involves three variables but has only one physical dimension, length ($\si{L}$). The fundamental dimensions can be combined in the form of products and/or quotients to generate derived dimensions such as area $\si{L}^2$ or, in other contexts, density $\si{M} \si{L}^{-3}$; \cite{sonin2001} provided more examples. Furthermore, ``dimension'' is related to but not quite the same as ``unit'' of measurement,
e.g., the meter as a measure of length; the distinction will be clarified shortly. 

In essence, the $\Pi$ theorem of \cite{buckingham1914} says that a physical system with $d$ variables and $p$ fundamental dimensions can be represented by $d - p$ dimensionless quantities. The theorem is implemented by choosing $p$ variables as {\em basis quantities\/} that represent all $p$ fundamental dimensions and correcting the dimensions of the remaining variables via the basis quantities.
In (\ref{eqn:pythag}), for instance, there are $d = 3$ variables and $p = 1$ for the dimension $\si{L}$. Arbitrarily, let $x_1$ be the basis quantity representing $\si{L}$, whereupon the remaining variables $y$ and $x_2$ can be turned into the dimensionless quantities
$q_0 = y / x_1$ and $q_1 = x_2 / x_1$. Equivalently, for the rest of the article, $d$ counts only the number of input variables: on the right of (\ref{eqn:pythag}) now $d = 2$ and there is $d - p = 1$ new input variable $q_1$ in addition to the output $q_0$. Throughout, derived DA variables will be denoted by $q$, with subscript 0 for the new output and subscripts $\jrange{1}{d-p}$ for the new inputs. Incidentally, the ``$\Pi$'' in the name of the theorem is only because the new variables were labelled $\Pi$, versus our $q$.

In general there are two constraints on the $p$ basis quantities. First, {\em representativity\/} says they must be chosen so that the dimensions of the remaining $d - p$ variables can be formed by combinations of the dimensions of the basis quantities. Secondly, {\em independence\/} (not statistical independence!) requires that the dimensions of a given basis quantity cannot be expressed in terms of combinations of the dimensions of the other basis quantities \citep{bridgman1931}. These conditions ensure that the remaining $d - p$ variables can be turned into new variables that are dimensionless. \cite{meinsma2019} gave two proofs that these new variables are sufficient to represent the system. In our simple illustration, (\ref{eqn:pythag}) can be written using only the two variables $q_0$ and $q_1$ as
\begin{equation}
\label{eqn:pythag:q}
q_0 = \frac{y}{x_1}  = \sqrt{\frac{x_1^2}{x_1^2} + \frac{x_2^2}{x_1^2}} = \sqrt{1 + q_1^2}.
\end{equation}
We also see from this simple example that DA leads to dimension reduction in the statistical sense: $d = 2$ inputs become a single input.

As long as units are consistent, e.g., $y$ and $x_1$ lengths are both measured in meters, the units cancel in dimensionless quantities; otherwise, only constants such as \si{m} / \si{ft} for length are introduced.

As already mentioned, complex computer codes cannot be quickly evaluated like (\ref{eqn:pythag}) or (\ref{eqn:pythag:q}) and a statistical surrogate is often employed. Accurate extrapolation is challenging for many statistical models, however, and even (\ref{eqn:pythag}) illustrates the challenge for a statistical model using the original variables $x_1$ and $x_2$. Suppose training data are collected where the lengths $x_1$ and $x_2$ are of the order centimeters and we try to extrapolate a predictive model to $x_1$ and $x_2$ of the order kilometers. A low-order polynomial regression or a GaSP model would both predict poorly so far outside the training ranges. \cite{AlbNacAlb2013} point out, however, that models using the dimensionless variables of DA
might scale much better. For our simple example with an empirical model relating $q_0$ to $q_1$, a triangle with $x_1 = 5 \si{cm}$ and $x_2 = 3 \si{cm}$ in the training data behaves just like a triangle with $x_1 = 5 \si{km}$ and $x_2 = 3 \si{km}$ in the test data: in the transformed space both have $q_1 = 3/5$ and there is no extrapolation. For given $x_1$ and $x_2$, a prediction $\hat{q}_0$ can be transformed back to the original scale as $\hat{y} = \hat{q}_0 x_1$. The examples we present for more complex applications demonstrate that large gains in prediction
accuracy (on the original scale) are possible from GaSP models even when the original input variables are substantially extrapolated outside their training ranges.

\section{Design and Analysis of Computer Experiments}
\label{sec:modeling}

We briefly review design and analysis of computer experiments, with emphasis on aspects that relate directly to our DA implementation.

The formulation and basic GaSP model go back to \cite{currin1991} and \cite{sacks1989}. The inputs to a computer code are denoted by the vector $\xVec = (\subs{x}{1}{d})^{\top} \in \mathbb{R}^d$ and the scalar output by $y = y(\xVec)$. The training design is $n$ input vectors $\xVec^{(1)}, \ldots, \xVec^{(n)}$, with corresponding output values in the vector $\mathbf{y} = (y_1, \ldots, y_n)^{\top}$. All examples in this article use maximin Latin hypercube designs \citep[mLHDs,][]{morris1995} for training and random Latin hypercubes for a test set to assess prediction; both types of design are implemented in the \texttt{R} library \texttt{DoE.wrapper} \citep{DoE.wrapper}. Furthermore, all designs are constructed in terms of the original variables. Following \cite{loeppky2009}, training designs will be of size $n = 10d$, with larger sizes also tried in some examples to investigate the rate of convergence of prediction error to zero.

A typical GaSP model is formulated in terms of the original variables 
$\xVec$ and $y(\xVec)$ of the computer code. Often in this article, we will use transformed variables from DA, but the GaSP formulation is the same: just replace $y$ by $q_0$ and $\xVec$ by $\subs{q}{1}{d-p}$ in the following (recall $p$ is the number of fundamental dimensions.) Essentially, this is just data pre-processing followed by the same methodology.

A GaSP model treats a deterministic output $y(\xVec)$ as a realization of a stochastic process,
\begin{equation}
\label{eq:gen_model}
Y(\xVec) = \mu(\xVec) + Z(\xVec),
\end{equation}
where $\mu(\xVec)$ is a regression function for the mean, and $Z(\xVec)$ is a correlated process with Gaussian marginal distribution $\mathcal{N}\left(0, \sigma^2 \right)$. We will be considering two alternatives for the regression component: $\mu(\xVec) = \beta_0$ or $\mu(\xVec) = \beta_0 + \sum_{j = 1}^d \beta_j x_j$. In both cases $\beta_0$ and the $\beta_j$ are unknown parameters to be estimated. We include the first-order regression model to investigate whether it can help with extrapolation. 

The correlation $R(\xVec, \xVec')$ between $Z(\xVec)$ and $Z(\xVec')$  at two input vectors $\xVec$ and $\xVec'$ is critical in a GaSP model. We will employ the power-exponential correlation function, with different scale and smoothness parameters for the $d$ inputs, but the methods proposed also apply if the user chooses, say, the Mat\'{e}rn class
\citep{stein1999}. Both will often lead to roughly equal prediction accuracy if the Mat\'{e}rn's smoothness parameters are also estimated data adaptively \citep{chen2016}. Whatever the choice, an important restriction is that the correlation function is a product of univariate correlation functions $R(\xVec, \xVec') = \prod_{j = 1}^d R_j(x_j, x'_j)$, to enable the sensitivity analysis central to our methodology for DA and described below. The product form is employed widely in the literature. 

All parameters of the GaSP model, i.e., $\beta_0$ (and optionally $\subs{\beta}{1}{d}$), $\sigma^2$, and the parameters of the chosen family of correlation functions will be estimated by maximum likelihood \citep{currin1991,sacks1989}.

\subsection{Prediction accuracy assessment}

Conditional on the trained model parameters, predictions at a test set of $N$ further input vectors $\xVec_{t}^{(1)}, \ldots, \xVec_{t}^{(N)}$
are from the best linear unbiased predictor \citep{sacks1989}  or equivalently the posterior mean of the process \citep{currin1991}.

Prediction accuracy is evaluated for all modelling approaches by  the test-set normalized root mean squared error (N-RMSE) expressed as a percentage:
\begin{equation}
\label{eq:n_rmse}
e_{\mathrm{N-RMSE}} = \frac{\sqrt{\frac{1}{N} \sum_{i = 1}^N \Big[ \hat{y}\left(\xVec_t^{(i)}\right) - y\left(\xVec_t^{(i)} \right) \Big]^2}}{\sqrt{\frac{1}{N} \sum_{i = 1}^N \Big[ \bar{y} - y\left(\xVec_t^{(i)} \right) \Big]^2}} \times 100\%,
\end{equation}
where $\bar{y} = (1/n) \sum_{i = 1}^n y_i$ is the mean output in the training data. The denominator is the test root mean squared error (RMSE) of the trivial predictor $\bar{y}$ and sets $e_{\mathrm{N-RMSE}}$ on a range of  0 to roughly 100\%, regardless of the scale of the original response. A value of 100\% indicates no better performance than the trivial predictor. All such metrics are computed on the original output scale, transforming back from the derived output variable in the case of DA. 

\subsection{Sensitivity analysis via FANOVA}

Our choice of basis quantities for DA (Section~\ref{sec:implementation}) is driven by a FANOVA sensitivity analysis. FANOVA decomposes the total variance of the GaSP predictor over the training domain into contributions from each individual input variable (its main effect) and from each pair of input variables (its 2-input interaction effect), plus higher-order terms. The percentage contributions attributed to these terms estimate their relative importances. We employ the FANOVA implementation described by \cite{schonlau2006} and implemented in {\tt R} by \cite{gasp}.

\section{Implementation of DA for a computer experiment}\label{sec:implementation}

\subsection{DA in four steps}

\cite{AlbNacAlb2013} and \cite{shen2014} described the implementation of Buckingham's $\Pi$ theorem in some detail. Briefly, but with some commentary about the novel implementations of the current article, there are four steps. 

\begin{enumerate}[label=\bfseries(\Roman*)] 

\item \textbf{The system and its dimensions.} 
For DA, we need to specify the respective fundamental dimensions of the $d$ inputs $\xVec$ and the output $y$. To generate a training design, the input ranges will be required; the ranges are also essential for our FANOVA-based determination of the basis quantities in the next step.

\item \label{basis} \textbf{Basis quantities.}
With a total of $p$ fundamental dimensions, choose $d - p$ inputs as basis quantities. They must satisfy the representativity and independence requirements mentioned in Section~\ref{sec:buckingham}. There will often be much latitude in this key step, and the current article advocates an empirical FANOVA approach.

\item \textbf{Dimensionless inputs and output.} 
Each of the remaining $d-p$ inputs is made dimensionless by transforming it using one or more of the basis quantities to create new inputs $\subs{q}{1}{d-p}$. Similarly, applying the basis quantities to $y$ generates a dimensionless output $q_0$. Choosing these transformations may again have many options, and we will employ FANOVA's visualization for a more complex application in Section~\ref{sec:solid_sphere}.   

\item \textbf{System's transformed function.} 
The fundamental result of Buckingham's $\Pi$ theorem says that the original specification $y(\xVec)$ 
can be represented as $q_0(\subs{q}{1}{d-p})$, as defined by the above steps,
and a GaSP model can use these variables.

\end{enumerate}

In Step~\ref{basis}, we advocate choosing as basis quantities the input variables with the most important contributions to predicted output variability (subject to the constraints of the DA theory). Our reasoning is explained by the following trivial example. A physical system follows $y = x_1^{1 - \alpha} x_2^{\alpha}$, where $x_1$, $x_2$ and $y$ have a single common physical dimension, and hence for dimensional consistency the exponents on the right sum to 1. Furthermore, suppose $\alpha \simeq 0$. Unless $x_2$ has a much larger range than $x_1$, we have $y \simeq x_1$ in the data; i.e., $x_1$ has the dominant effect. Choosing the important factor $x_1$ as the basis quantity and modeling $q_0 = y / x_1$ as a function of $q_1 = x_2 / x_1$ turns the true physical system into $q_0 = y / x_1 = x_1^{1 - \alpha} x_2^{\alpha} / x_1 = x_2^\alpha / x_1^\alpha = q_1^\alpha$, which is approximately constant and easy to model because $\alpha \simeq 0$. In contrast, choosing the unimportant factor $x_2$ as the basis quantity, with $q_0 = y / x_2$ and $q_1 = x_1 / x_2$, gives $q_0 = y / x_2 = x_1^{1 - \alpha} x_2^{\alpha} / x_2 = (x_1 / x_2)^{1-\alpha} = q_1^{1-\alpha}$. The approximately linear relation ($\alpha \simeq 0$) is still easy to model in this simple example, but not as easy as a constant. Note that $q_0$ now has variability due to $y$ {\em and\/} $x_2$,  when $x_2$ was originally unimportant, which must be modelled via a $q_1$ effect. The same intuition might apply to a more realistic physical system: correcting the dimensions of $y$ using unimportant variables just introduces more variation to be modelled. This intuition is borne out by the empirical studies of this article, where the DA choices leading to modeling $q_0(\subs{q}{1}{d-p})$ give more accuracy than using the raw variables (and some choices lead to less accuracy).

\subsection{Simple example} 
A simple example will illustrate the above basic steps in DA and the proposed FANOVA methodology to guide those steps.

\begin{enumerate}[label=\bfseries(\Roman*)] 

\item \textbf{The system and its dimensions.} 
The vertical displacement $y$ of an object falling due to gravity in a vacuum as a function of $d = 4$ inputs is
\begin{equation}
\label{eq:displacement_function}
y = y_0 + V_0 t - \frac{g t^2}{2},
\end{equation}
where the output is displacement $y$ with dimension $[y] = \si{L}$ in m. The dimensions of the four inputs are in Table~\ref{tab:gravity_displacement_inputs}, along with their units and ranges.

\begin{table}[ht!]
\begin{center}
\caption{Inputs, dimensions, units, and ranges for the gravity displacement function.}
\begin{tabular}{lllcc}
\hline
&&& \multicolumn{2}{c}{\textbf{Range}} \\
\cline{4-5}
\textbf{Input} & \textbf{Dimensions} & \textbf{Units} & \textbf{Training} & \textbf{Extrapolation} \\ 
\hline
$y_0$, initial position & $\si{L}$             & $\si{m}$             & $[1, 10]$ & $[10, 20]$ \\
$V_0$, initial velocity & $\si{L} \si{T}^{-1}$ & $\si{m} \si{s}^{-1}$ & $[1, 10]$ & $[10, 20]$   \\ 
$t$, time of displacement & $\si{T}$               & $\si{s}$        & $[1, 10]$  &  $[10, 20]$    \\ 
$g$, gravitational acceleration & $\si{L} \si{T}^{-2}$ & $\si{m} \si{s}^{-2}$ & $[1.62, 9.81]$ & $[10.44, 24.79]$    \\ \hline
\end{tabular}
\label{tab:gravity_displacement_inputs}
\end{center}
\end{table}

We will treat the function as unknown to show the consequences of various DA implementation decisions and the resulting GaSP models. Furthermore, the training study will allow environments where the gravitational acceleration $g$ varies between that on the moon and on the earth, so all $d = 4$ inputs are active. 

\item \textbf{Basis quantities.} 
As Table~\ref{tab:gravity_displacement_inputs} has $p = 2$ fundamental dimensions---length ($\si{L}$) and time ($\si{T}$)---we need to choose two basis quantities. The representativity and independence constraints allow many choices even in this simple example, and the $\Pi$ theorem gives no guidance.

\begin{itemize}

\item
{\bf Initial DA.}  
To represent $\si{L}$ and $\si{T}$, a straightforward choice of basis quantities is $y_0$ and $t$, with dimensions $\si{L}$ and $\si{T}$, respectively.

\item
{\bf FANOVA DA.}
To identify another set of basis quantities for DA, we use FANOVA to identify the most important inputs. We train a GaSP model with a constant regression term and the squared exponential correlation function. Throughout this article all models for a given example are trained using a common set of data in the original variables. 

\begin{center}
\begin{figure}[ht!]
\centering
    \includegraphics[width = 0.68\textwidth]{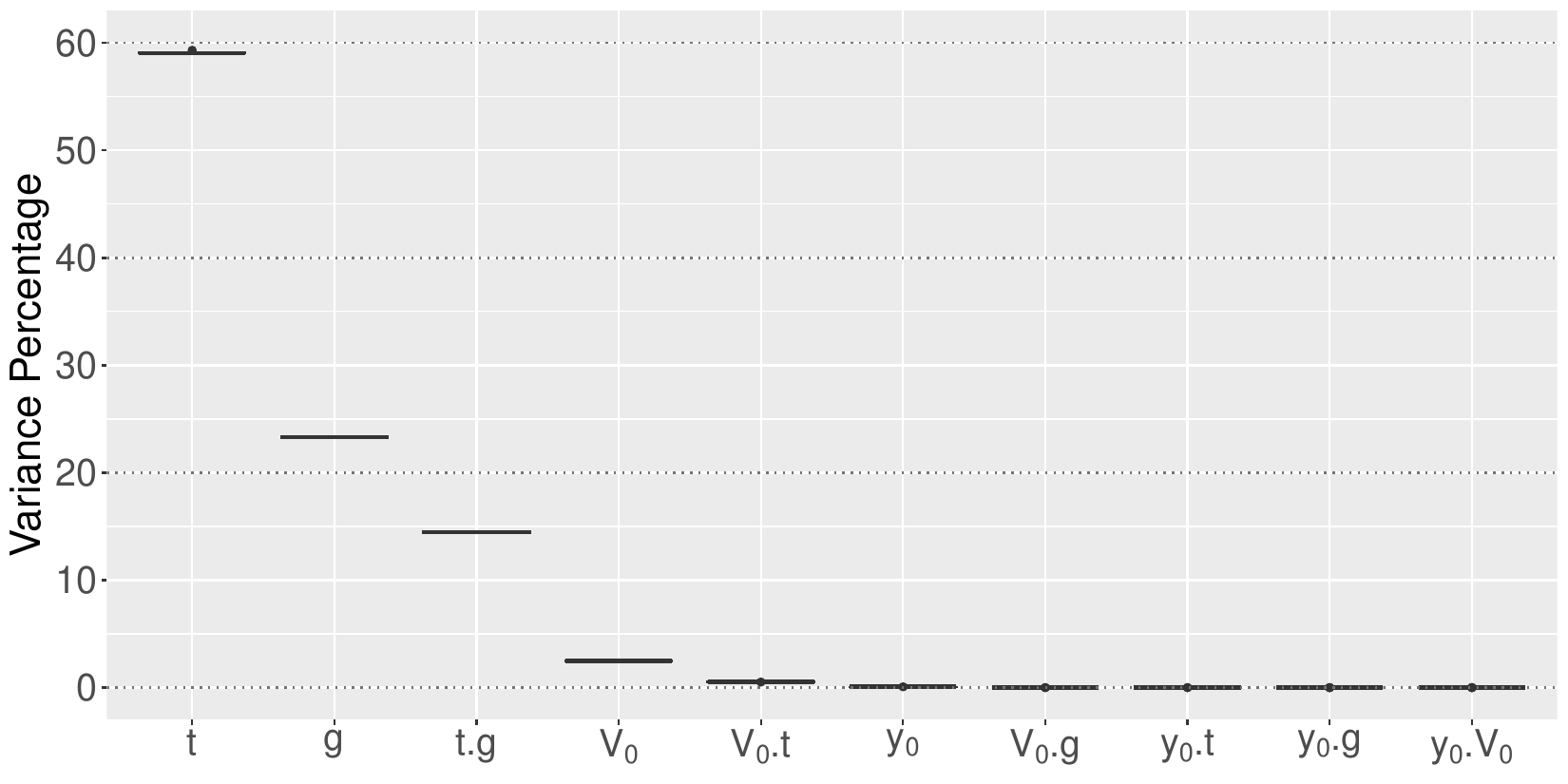}
\caption{FANOVA percentage contributions for the gravity displacement function using a GaSP model with squared exponential correlation function and a constant regression term. Each boxplot shows results from 20 mLHDs of $n = 20$ runs for training.}
\label{fig:gravity_displacement_FANOVA}
\end{figure}
\end{center}

We do not create novel designs tailored to variables identified by DA 
\citep{shen2018_2}. The training data are from an mLHD of size $n = 20$ in the four original inputs, and the experiment is repeated 20 times just to demonstrate that the FANOVA conclusions generalize. Integration in FANOVA is with respect to uniform weights over the input ranges in Table~\ref{tab:gravity_displacement_inputs}; uniform weights will be used in all examples. Recall that FANOVA decomposes the total variance of a GaSP predictor into contributions from main effects, 2-input interactions, etc. 

The FANOVA contributions here, as computed by the \texttt{GaSP} package \citep{gasp}, are plotted in Figure~\ref{fig:gravity_displacement_FANOVA}. We see that $t$ and $g$ and their interaction consistently provide the largest percentage contributions  to the variance of the predictor, and FANOVA DA uses them as basis quantities.

\end{itemize}

\item \textbf{Dimensionless inputs and output.}
The two basis quantities are applied to the output and the other two (non-basis) inputs, to make them  dimensionless.

\begin{itemize}

\item
{\bf Initial DA.}  
The basis quantities $y_0$ and $t$ have dimensions $\si{L}$ and $\si{T}$, respectively. The output $y$ has its dimension $\si{L}$ eliminated by dividing by $y_0$:
\begin{equation}
\label{eq:output_gravity_initial_DA}
q_0^{(I)} = \frac{y}{y_0},            
\end{equation}
with superscript $I$ signifying initial DA.
Similarly, the inputs that are not basis quantities, $V_0$ and $g$, are made dimensionless:
\begin{equation}
\label{eq:inputs_gravity_initial_DA}
q_1^{(I)} =\frac{V_0 t}{y_0} \qquad \text{and} \qquad q_2^{(I)} = \frac{g t^2}{y_0}.
\end{equation}
Note that $V_0$ and $g$ each involve $\si{L}$ and $\si{T}$, 
and both new inputs have to be corrected using both basis quantities.
 
\item
{\bf FANOVA DA.}
Applying $t$ and $g$ to correct the other original variables, the FANOVA DA has the dimensionless output
\begin{equation}
\label{eq:output_gravity_FANOVA_DA}
q_0^{(F)} = \frac{y}{g t^2}
\end{equation}
and two dimensionless inputs
\begin{equation}
\label{eq:inputs_gravity_FANOVA_DA}
q_1^{(F)} = \frac{y_0}{g t^2} \qquad \text{and} \qquad q_2^{(F)} = \frac{V_0}{g t},
\end{equation} 
with superscript $F$ for FANOVA.

\end{itemize}

\item {\bf System's transformed function.}
Thus, we either fit a GaSP with output $q_0^{(I)}$ and inputs $q_1^{(I)}$ and $q_2^{(I)}$
(initial DA) 
or with output $q_0^{(F)}$ and inputs $q_1^{(F)}$ and $q_2^{(F)}$ (FANOVA DA). 

\end{enumerate}

\begin{center}
\begin{figure}[ht!]
\centering
    \includegraphics[width = 0.75\textwidth]{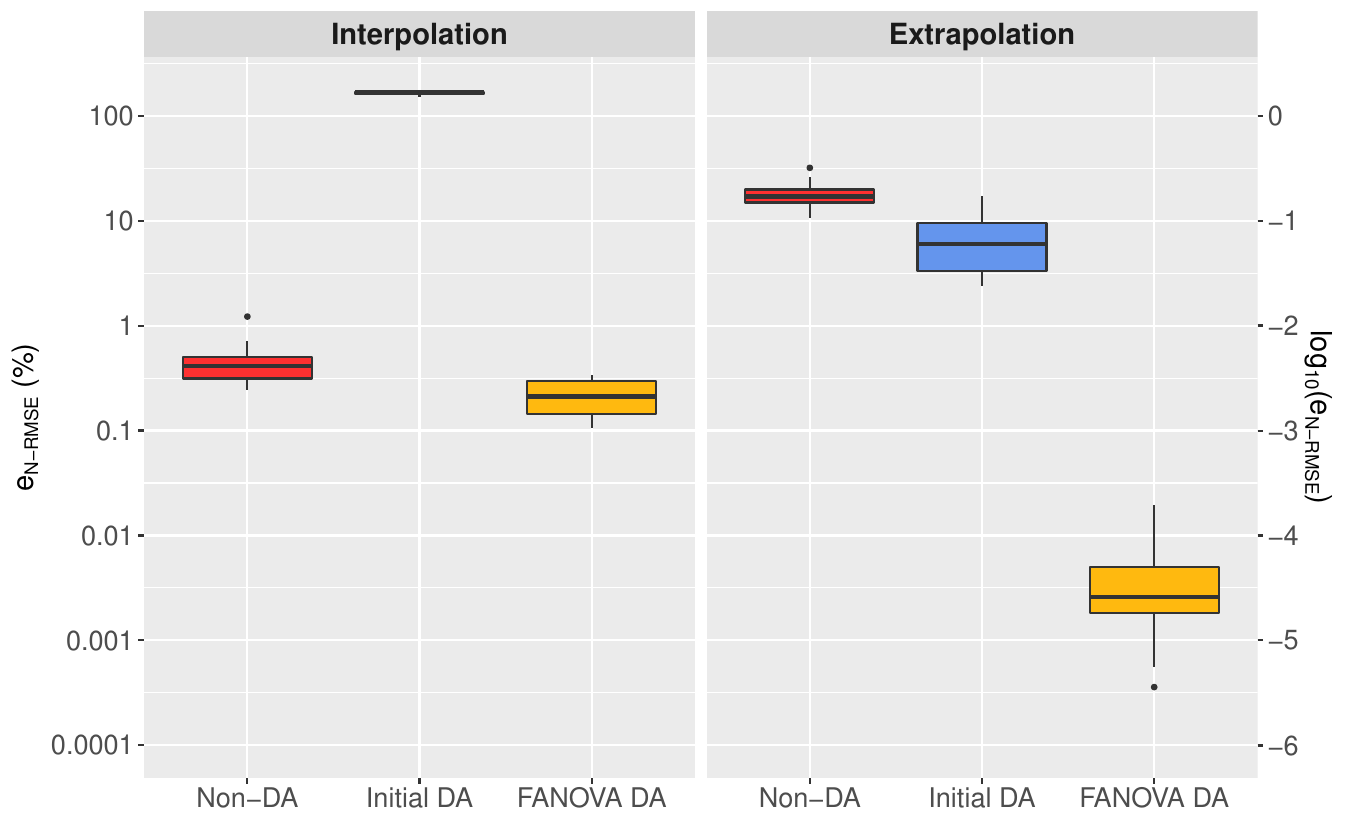}
    \caption{Prediction accuracy by type of DA for the gravity displacement function using a GaSP model with squared exponential correlation function and a constant regression term. Each boxplot shows results from 20 mLHDs of $n = 20$ runs for training and $N = 10,000$ runs for testing.}
\label{fig:gravity_displacement_N_RMSE}
\end{figure}
\vspace{-8mm}
\end{center}

We compare the prediction accuracy of three different approaches: non-DA (original inputs and output), initial DA, and FANOVA DA. Twenty repeat experiments use mLHDs of size $n = 40$ in the original inputs; these data are transformed to form the two sets of quantities in the respective DA models.

The interpolation and extrapolation assessments use two different test sets of size $N = 10,000$ points from random LHDs. For interpolation the test LHD uses the  training ranges in Table~\ref{tab:gravity_displacement_inputs}. The extrapolation ranges, also shown in Table~\ref{tab:gravity_displacement_inputs}, are outside the training ranges for every variable. In particular, $g$ is extrapolated to $[10.44, 24.79]$; these limits are the  gravitational forces on Saturn and Jupiter, respectively. All accuracy assessments via the normalized RMSE in~(\ref{eq:n_rmse}) are on the original scale: the DA output $q_0^{(I)}$ or $q_0^{(F)}$ is transformed back to $y$ before computing errors.

Figure~\ref{fig:gravity_displacement_N_RMSE} depicts the interpolation and extrapolation prediction accuracy from the three strategies. The boxplots show the variation in the assessment metric $e_{\mathrm{N-RMSE}}$ over the 20 repeat training experiments. The non-DA model achieves high accuracy (nearly always less than 1\% normalized test RMSE) for interpolation but is poor for extrapolation. The initial DA fares even worse, showing little or no predictive power for either interpolation or extrapolation, highlighting that an arbitrary implementation of DA, while it satisfies Buckingham's theorem, can be detrimental.  The FANOVA DA, however, achieves even higher accuracy than non-DA and remarkably performs even better for extrapolation, achieving a near-perfect predictor.    

\section{Borehole Function}
\label{sec:b_function}

The borehole function is a well known test problem in the computer experiments literature going back to \cite{worley1987} and \cite{morris1993}. Like the illustrative gravitational example, it is a fast, algebraic function allowing multiple training sets to establish generalization beyond a single experiment and a large test set. 

The borehole function models the flow of water through a drilled borehole from the ground surface to two aquifers. The output $y_\mathrm{b}$, flow rate through the borehole in \si{m^3 year^{-1}} 
(i.e., $[y_\mathrm{b}] = \si{L}^3 \si{T}^{-1}$), is a function of eight input parameters:
\begin{equation}
\label{eq:b_function}
y_\mathrm{b}(\xVec) = \frac{2 \pi T_u(H_u - H_l)}{\log (r / r_w) \Big[1 + \frac{2 L T_u}{\log(r / r_w) r^2_w K_w} + \frac{T_u}{T_l} \Big]};
\end{equation}
where the inputs and their respective dimensions, units, and ranges are detailed in Table~\ref{tab:borehole:inputs}. Previous studies show that the borehole function is predicted well by a GaSP with a modest training size when the purpose is interpolation, but we will see that extrapolation again poses challenges unless a careful DA is implemented.
 
\begin{table}[ht!]
\begin{center}
\caption{Inputs, dimensions, units, and ranges for the borehole function.}
\begin{tabular}{lllcc}
\hline
&&& \multicolumn{2}{c}{\textbf{Range}} \\
\cline{4-5}
\textbf{Input} & \textbf{Dimensions} & \textbf{Units} & \textbf{Training} & \textbf{Extrapolation} \\ 
\hline
$r_w$, radius of borehole & $\si{L}$               & $\si{m}$        & $[0.05, 0.15]$ &   $[0.15, 0.25]$ \\
$r$, radius of influence & $\si{L}$               & $\si{m}$        & 
\multicolumn{2}{c}{$[100, 50000]$}    \\ 
\begin{tabular}[c]{@{}c@{}}$T_u$, transmissivity \\ of upper aquifer\end{tabular} & $\si{L}^2 \si{T}^{-1}$ & $\si{m^2 year^{-1}}$ & \multicolumn{2}{c}{$[63070, 115600]$} \\
\begin{tabular}[c]{@{}c@{}}$H_u$, potentiometric head\\ of upper aquifer\end{tabular} & $\si{L}$               & $\si{m}$        & $[990, 1110]$ & $[1110, 1170]$   \\ 
\begin{tabular}[c]{@{}c@{}}$T_l$, transmissivity \\ of lower aquifer\end{tabular} & $\si{L}^2 \si{T}^{-1}$ & $\si{m^2 year^{-1}}$ & \multicolumn{2}{c}{$[63.1, 116]$} \\
\begin{tabular}[c]{@{}c@{}}$H_l$, potentiometric head\\ of lower aquifer\end{tabular} & $\si{L}$               & $\si{m}$        & $[700, 820]$ & $[820, 880]$   \\ 
$L$, length of borehole & $\si{L}$               & $\si{m}$        & $[1120, 1680]$ & $[1680, 1960]$   \\ 
\begin{tabular}[c]{@{}c@{}}$K_w$, hydraulic conductivity \\ of borehole\end{tabular} & $\si{L } \si{T}^{-1}$  & $\si{m \text{ }year^{-1}}$ & \multicolumn{2}{c}{$[9855, 12045]$}   \\
\hline
\end{tabular}
\label{tab:borehole:inputs}
\end{center}
\end{table}

We will see that prediction accuracy is also improved by appropriate input transformations and an input space expansion, with DA providing extra benefit. The non-DA and DA methods all have a no-transformation baseline for comparison: the inputs and output in the GaSP model are the original variables in~(\ref{eq:b_function}) for non-DA, or the dimensionless quantities derived from them for DA. Then we adapt these baseline models with logarithmic transformations of their respective inputs, as well as an input-space expansion. 
Further details about these transformations will appear in Section~\ref{sect:borehole:settings}. The results for this progression of models will show the advantages of working on the logarithmic scale for both non-DA and DA frameworks.

Again, all training and test designs in the original variables are common to all methods, with logarithmic or DA transformations applied to common data. 

\subsection{Dimensional analysis}

\begin{figure}[ht!]
\begin{center}
\begin{subfigure}{1\textwidth}
\begin{center}
\includegraphics[width=0.8\linewidth]{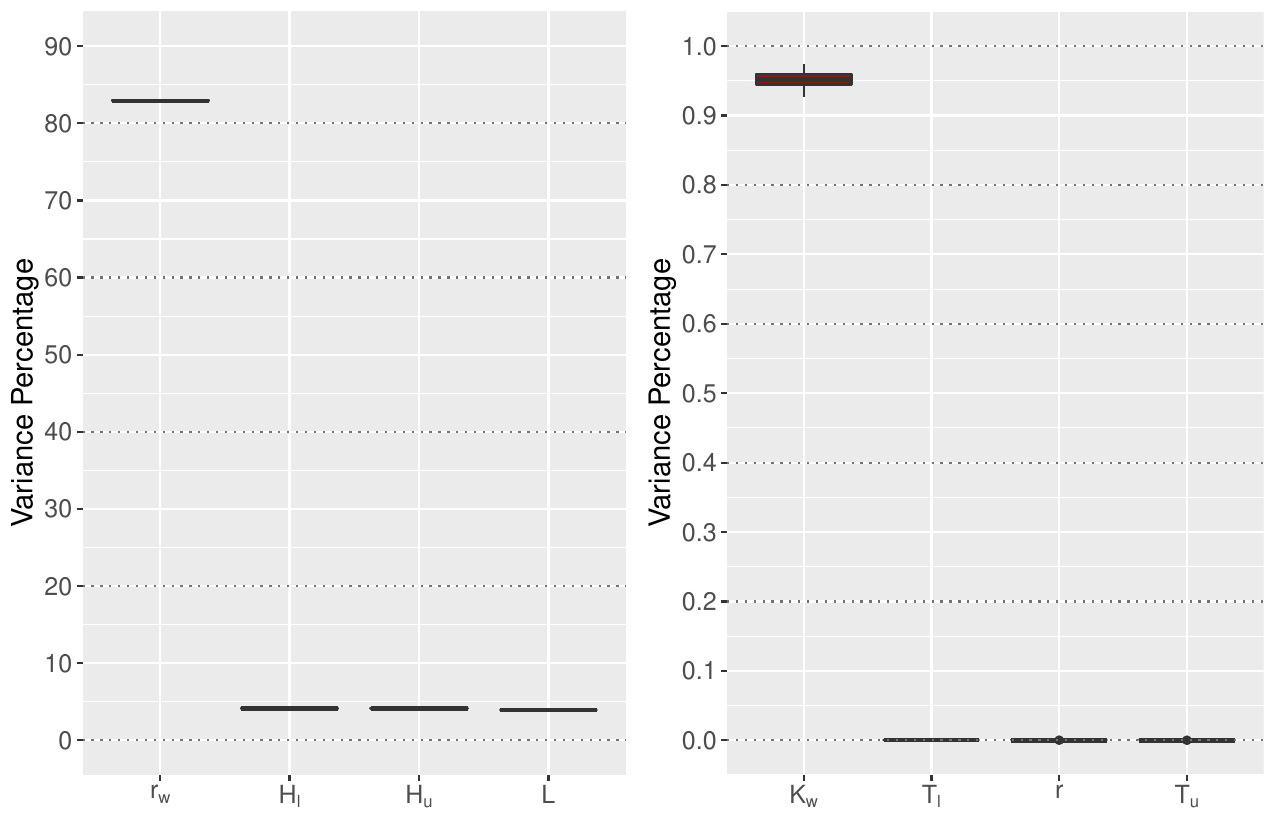}
\vspace{-0.4\baselineskip}  
\caption{Main effects}
\label{fig:b_function_main_effects_plot}
\end{center}
\end{subfigure}
\vspace{1\baselineskip}  
\begin{subfigure}{1\textwidth}
\begin{center}
\includegraphics[width=0.8\linewidth]{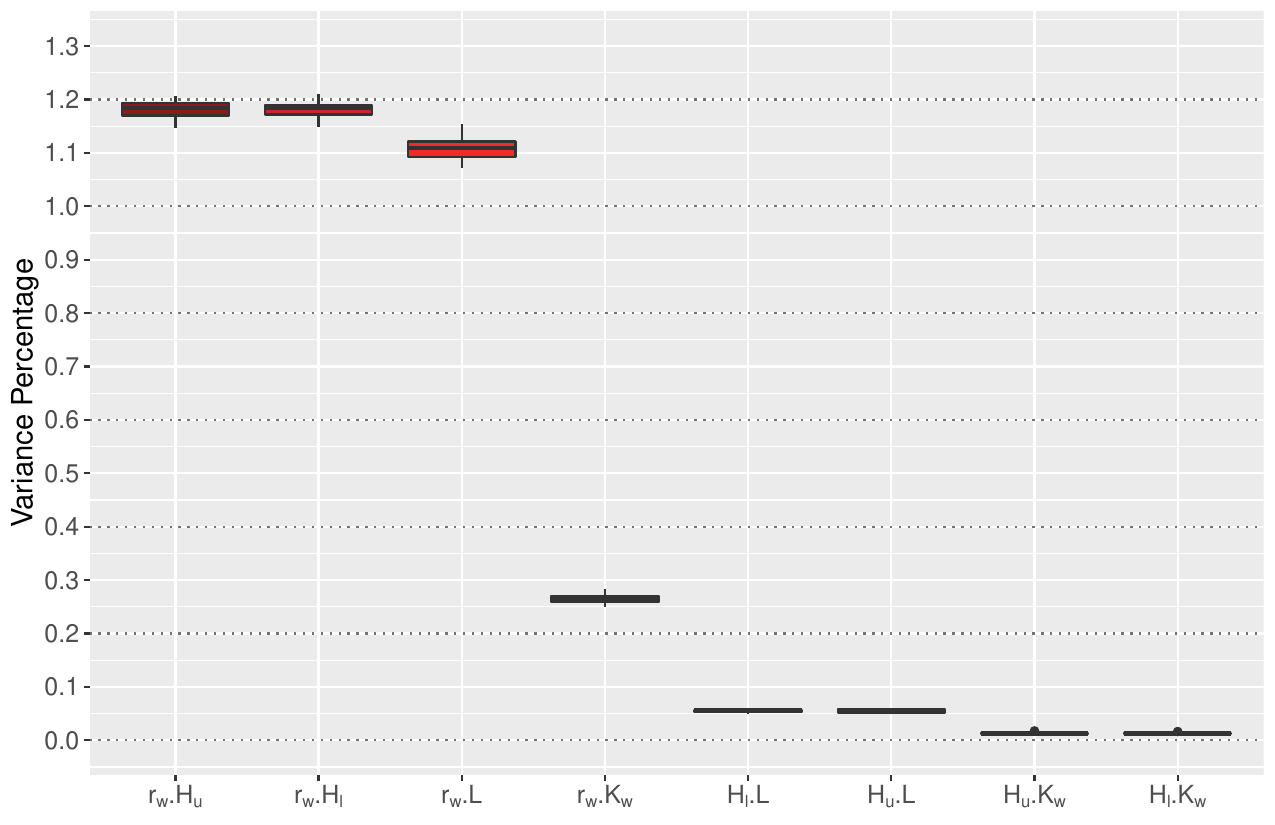}
\vspace{-0.4\baselineskip} 
\caption{Interaction effects}
\label{fig:b_function_input_interactions_plot}
\end{center}
\end{subfigure}
\vspace{-1.3\baselineskip}
\caption{FANOVA percentage contributions for the borehole function using a GaSP model with power exponential correlation function and a constant regression term. Each boxplot shows results over 20 repeat experiments
with different mLHDs of $n = 80$ runs for training.
(a) Main effects, with larger and smaller percentages in two panels with different scales.
(b) The largest 2-input interaction contributions.}
\label{fig:b_function_FANOVA}
\end{center}
\end{figure}

\begin{enumerate}[label=\bfseries(\Roman*)] 

\item \textbf{The system and its dimensions.} 
The fundamental dimensions in Table~\ref{tab:borehole:inputs} are length ($\si{L}$) and time ($\si{T}$), i.e., $p = 2$.

\item \textbf{Basis quantities.} 
Since there are $p = 2$ fundamental dimensions, we need to select two basis quantities for DA.

\begin{itemize}

\item
{\bf FANOVA DA.} 
FANOVA therefore needs to identify the two most important inputs satisfying the representativity and independence conditions. We run FANOVA for 20 repeat experiments with training size of $n = 10d = 80$ runs, the power exponential correlation function, and a constant regression term. Here the original inputs in Table~\ref{tab:borehole:inputs} (on their training ranges) and the original output $y_\mathrm{b}$ are used in a GaSP model. We will see that $n = 80$, the smallest training size considered in later comparisons, leads to consistent FANOVA conclusions over all 20 repeat experiments.
 
\begin{itemize}

\item \textbf{Main effects.} 
Figure~\ref{fig:b_function_main_effects_plot} shows the respective variance percentages of the eight inputs across the 20 repeated experiments. Note there is little variability in percentage across the repeat experiments: a single experiment would give reliable conclusions. We see that the four most important main effects in the left panel---$r_w$, $H_u$, $H_l$, and $L$---all involve only the fundamental dimension length ($\si{L}$). However, the main effect of radius of borehole $r_w$ dominates, accounting for a median of 82.5\% of total variation. To find a second basis quantity, for time $\si{T}$, we turn to the inputs with smaller percentages in the right panel of Figure~\ref{fig:b_function_main_effects_plot}. Among them, hydraulic conductivity $K_w$ is the most important main effect involving $\si{T}$  (and \si{L}), with percentage contribution around 0.95\%.

\item \textbf{Input interactions.} 
Figure~\ref{fig:b_function_input_interactions_plot} shows the variance percentages for the top eight 2-input interactions. The most important are $r_w \cdot H_u$, $r_w \cdot H_l$, and $r_w \cdot L$, each accounting for just over 1\% of the total variance; they only involve fundamental dimension $\si{L}$. The fourth 2-input interaction, $r_w \cdot K_w$, has dimensions $\si{L}$ and $\si{T}$. Again, $K_w$ emerges to represent $T$. 

\end{itemize}

Since $r_w$ has a dominant role in FANOVA followed by $K_w$, we use them as basis quantities in FANOVA DA. 

\item 
\textbf{Shen, Lin, and Chang (2018) DA (SLC DA).}
These authors also used the borehole function as one of their examples. They selected potentiometric head of upper aquifer ($H_u$) and transmissivity of upper aquifer ($T_u$) as basis quantities, due to their smallest range ratios. We implement their DA for comparison.

\end{itemize}

\item \textbf{Dimensionless inputs and output.} 
Thus, the previous choices for the two DA frameworks 
lead to the following input and output transformations.

\begin{itemize}

\item \textbf{FANOVA DA.} 
The inputs $r_w$ and $K_w$ identified by FANOVA are used to make the other variables dimensionless. The new output is
\begin{equation}
\label{eq:output_DA_FANOVA}
q_0^{(F)} = \frac{y_\mathrm{b}(\xVec)}{r_w^2 K_w},
\end{equation}
and the $d - p = 8 - 2 = 6$ dimensionless inputs  $q_i^{(F)}$ for $i = 1, \ldots, 6$ are
\begin{equation}
\label{eq:inputs_DA_FANOVA}
\frac{r}{r_w}, \,       \frac{T_u}{r_w K_w}, \, \frac{H_u}{r_w}, \,
\frac{T_l}{r_w K_w}, \, \frac{H_l}{r_w}, \,     \frac{L}{r_w},
\end{equation}
respectively.

\item
\textbf{SLC DA.}
Taking instead the basis quantities $H_u$ and $T_u$, the new output is
\begin{equation}
\label{eq:output_DA_Shen}
q_0^{(S)} = \frac{y_\mathrm{b}(\xVec)}{H_u T_u},
\end{equation}
and the dimensionless inputs $q_i^{(S)}$ for $i = 1, \ldots, 6$ are
\begin{equation}
\label{eq:inputs_DA_Shen}
\frac{r_w}{H_u}, \, \frac{r}{H_u}, \, \frac{T_l}{T_u}, \, 
\frac{H_l}{H_u}, \, \frac{L}{H_u}, \, \frac{K_w H_u}{T_u},
\end{equation}
respectively.
\end{itemize}

\item \textbf{System's transformed function.}
The two DA representations are hence
\begin{equation*} 
\frac{y_\mathrm{b}(\xVec)}{r_w^2 K_w} = g \left( \frac{r}{r_w}, \frac{T_u}{r_w K_w}, \frac{H_u}{r_w}, \frac{T_l}{r_w K_w}, \frac{H_l}{r_w}, \frac{L}{r_w} \right)
\end{equation*}
for FANOVA DA, and
\begin{equation*}
 \frac{y_\mathrm{b}(\xVec)}{H_u T_u} = h \left( \frac{r_w}{H_u}, \frac{r}{H_u}, \frac{T_l}{T_u}, \frac{H_l}{H_u}, \frac{L}{H_u}, \frac{K_w H_u}{T_u} \right)
\end{equation*}
for SLC DA.
GaSP models are trained for either $g(\cdot)$ or $h(\cdot)$.

\end{enumerate}

\subsection{Simulation settings}\label{sect:borehole:settings}

In all simulations, we implement two possible regression components: constant or linear terms. Previous arguments \citep{chen2016} suggested linear trend terms are unnecessary, but we check whether this conclusion carries over to DA and whether inclusion of such terms is helpful for extrapolation. Hence, we compare the prediction accuracy of three different approaches, non-DA, FANOVA DA, and SLC DA, each with a constant or linear-trend regression model.

Furthermore, three types of input transformation are tried.

\begin{itemize}

\item \textbf{No input transformation.}
As a baseline the inputs of the respective models are used directly. 

\item \textbf{Log input.}
All inputs are transformed to their logs. In the non-DA models, taking logarithms of the original inputs is particularly relevant for the radius of influence ($r$), as it ranges over several orders of magnitude. In the FANOVA DA models, $q_1^{(F)}$ is replaced by $\ln(q_1^{(F)})$, for example, which is allowed in the DA theory because $q_1^{(F)}$ is dimensionless; logarithms are not applied to $r$ and $r_w$ within the ratio $r / r_w$ in $q_1^{(F)}$. Of course $\ln(r / r_w) = \ln(r) - \ln(r_w)$, but it is only this difference, not the individual logged inputs, that appear in the FANOVA DA model. For all approaches, log-transformed inputs are used in the regression model if it has linear terms as well as in the correlation function.

\item \textbf{Expanded log-input.} 
In addition to  the log-transformed inputs as above, there is a substantial expansion of the input space here: all sums and differences of the logged variables also appear in the correlation function. This yields $d = 8 + 2{{8}\choose{2}} = 64$ input variables for non-DA and $d = 6 + 2{{6}\choose{2}} = 36$ for FANOVA DA and SLC DA. While all these variables appear in the correlation function, a regression model with linear terms only includes terms in the original or log-transformed inputs, without expansion.

Logarithmic transformation and expansion generates new inputs, as we now illustrate. For non-DA, they include some (but not all) of the dimensionless DA inputs, for example, $\ln(r) - \ln(r_w)$ is $\ln(q_1^{(F)})$, i.e., there is an implicit partial DA. FANOVA DA and SLC DA potentially enjoy higher-order benefits. For instance, $\ln(q_1^{(F)}) + \ln( q_2^{(F)}) =  \ln((r / r_w) T_u / (r_w K_w)) = \ln(r T_u / K_w)$,
a new dimensionless quantity involving three of the original inputs. Even though the computing time for maximum likelihood optimization will increase with the number of inputs, the impact is not as large as an increase in the training set size $n$.

\end{itemize}

The $n = 10d$ rule of thumb gives $n = 80$ here, increasing to 160, 320, and 640 to investigate the effect of $n$ on accuracy. There are 20 different mLHDs per training size, and test set size is fixed at $N = 10,000$. 

\subsection{Interpolation}

\begin{figure}[ht!]
\centering
    \includegraphics[width = 1\textwidth]{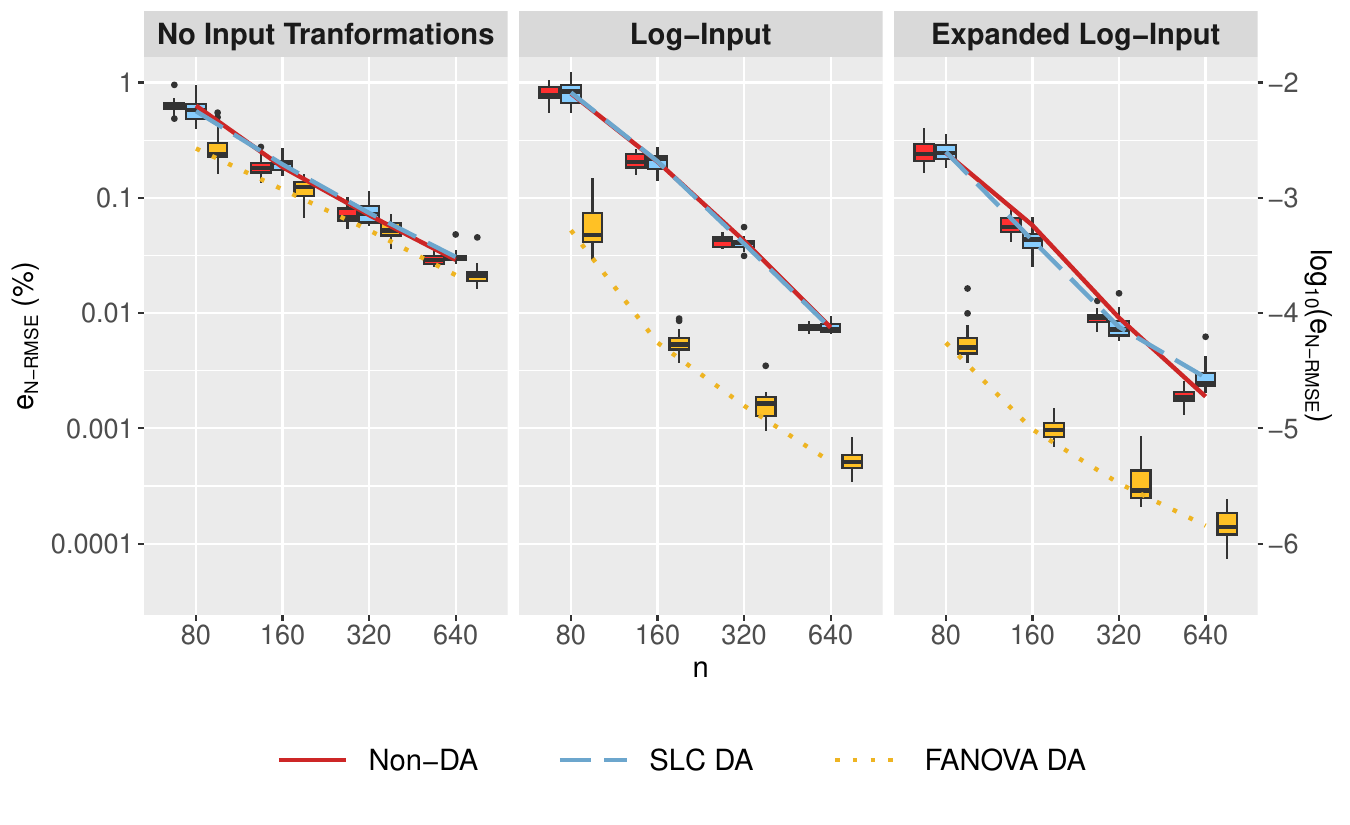}
    \caption{Interpolation prediction accuracy for the borehole function by type of DA
    and three input arrangements.  All models have a constant regression component. 
   Each boxplot shows results from 20 mLHDs of $n = 80, 160, 320, 640$ runs for training 
   and $N = 10,000$ runs for testing.
   The lines join $e_{\mathrm{N-RMSE}}$ sample means.}
\label{fig:borehole:interpolation}
\end{figure}

Figure~\ref{fig:borehole:interpolation} illustrates prediction accuracy as a function of sample size for all methods using a constant regression model. The three strategies non-DA, FANOVA DA, and SLC DA are compared in each panel, with the three possible input transformations (untransformed, log, expanded log) in the different panels. The boxplots show the variation in the assessment metric $e_{\mathrm{N-RMSE}}$ across the 20 repeat experiments from distinct mLHDs. To avoid overlap of the boxplots for a given sample size, they are offset horizontally; the offset is corrected for the lines joining the $e_{\mathrm{N-RMSE}}$ sample means.

The first panel of Figure~\ref{fig:borehole:interpolation} shows that FANOVA DA outperforms the other two approaches without any input transformation. The performance measure on the $y$-axis has a log scale, and the differences are more important than may be apparent. It is also seen in the next two panels that all three strategies benefit considerably from logarithmic transformation of the inputs and benefit further from the expanded-log inputs. FANOVA DA gains by far the most.
It achieves an average $e_{\mathrm{N-RMSE}}$ of 0.006\% at $n = 80$ and ends with 0.0002\% at $n = 640$. Hence this method achieves near-perfect accuracy even with $n = 80$. 

The two axes in Figure~\ref{fig:borehole:interpolation} are on logarithmic scales, and all lines joining the sample means of $e_{\mathrm{N-RMSE}}$
are approximately linear throughout the range, suggesting a constant rate of convergence to zero. An almost 10-fold increase in sample size leads to 1 to 2 orders of magnitude improvement in
$e_{\mathrm{N-RMSE}}$ (or RMSE), i.e., a rate of convergence to zero better than $n^{-1}$. (Note this is for root MSE, not MSE.)

As shown in Figure~\ref{fig:b_function_linear_interpolation_plot} of Appendix~\ref{sect:appendix}, the inclusion of linear regression terms does not improve prediction accuracy for any of the modelling strategies.

\subsection{Extrapolation}

We now turn to extrapolation. \cite{shen2018_2} argue that when input ranges are extrapolated a model satisfying dimensional constraints may perform better, especially for a complex response surface. FANOVA DA and SLC DA both satisfy the dimensional constraints and are compared with non-DA.

We obtain extrapolated predictions from the repeat experiments and trained models described above for interpolation, i.e., with training input ranges as in Table~\ref{tab:borehole:inputs}. For extrapolation, however, four inputs have extended ranges in a new test set of $N = 10,000$ point. The FANOVA in Figure~\ref{fig:b_function_main_effects_plot} indicated that radius of borehole ($r_w$), potentiometric head of lower aquifer ($H_l$), potentiometric head of upper aquifer ($H_u$), and length of borehole ($L$) are four important inputs, and the test set now has the extrapolated ranges for these variables in Table~\ref{tab:borehole:inputs}.

\begin{figure}
\centering
    \includegraphics[width = 1\textwidth]{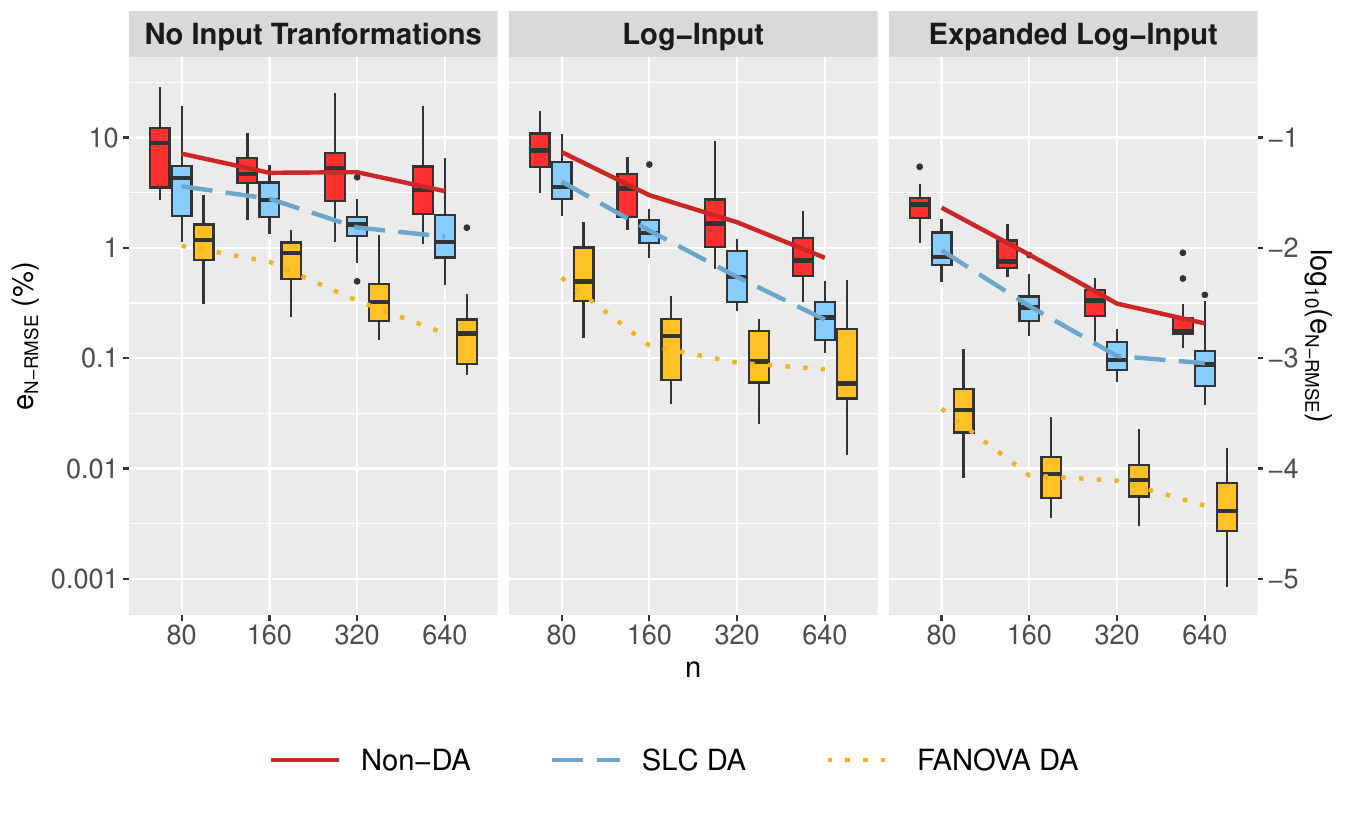}
    \caption{Extrapolation prediction accuracy for the borehole function 
   by type of DA and three input arrangements.  All models have a constant regression component. 
   Each boxplot shows results from 20 mLHDs of $n = 80, 160, 320, 640$ runs for training, 
   and the test set is is on the extrapolated ranges in Table~\ref{tab:borehole:inputs}.
   The lines join $e_{\mathrm{N-RMSE}}$ sample means.}
\label{fig:borehole:extrapolation}
\end{figure}

Figure~\ref{fig:borehole:extrapolation} shows prediction accuracy for extrapolation. Not surprisingly, for all strategies accuracy deteriorates slightly relative to interpolation. Again FANOVA DA outperforms the other two approaches, with or without input transformation. In conjunction with the expanded log-inputs, it still achieves an average $e_{\mathrm{N-RMSE}}$ of about 0.032\% at $n = 80$ and 0.0032\% at $n = 640$, i.e., high accuracy for the smallest sample size and excellent accuracy for the largest. In contrast, the other two methods benefit from input expansion but only achieve average $e_{\mathrm{N-RMSE}}$ around 1\% or worse at $n = 80$, with SLC DA outperforming non-DA.

Furthermore, as shown in Figure~\ref{fig:b_function_linear_extrapolation_plot} of Appendix~\ref{sect:appendix}, use of a linear regression component does not improve prediction accuracy compared to the constant regression. Linear trend terms do not aid extrapolation here.

\newpage
\section{Heat Transfer in a Solid Sphere}
\label{sec:solid_sphere}

The second case study was used by \cite{tan2017} to illustrate DA for polynomial emulation of a complex computer code. Thus the methodology was not aimed at GaSP models, but the example is useful to illustrate DA in a more complex setting.

\begin{table}[ht!]
\begin{center}
\caption{Inputs, dimensions, units, and ranges for the solid-sphere heat transfer function.}
\begin{tabular}{lllcc}
\hline
&&& \multicolumn{2}{c}{\textbf{Range}} \\
\cline{4-5}
\textbf{Input} & \textbf{Dimensions} & \textbf{Units} & \textbf{Training} & \textbf{Extrapolation} \\ 
\hline
\begin{tabular}[c]{@{}c@{}}$R$, ratio of distance from \\ center and sphere radius\end{tabular}              & Dimensionless                    & None                          & \multicolumn{2}{c}{$[0.01, 1]$}   \\
$r$, radius of sphere                                                                                       & $\si{L}$                         & \si{m}                      & $[0.05, 0.2]$ &  $[0.2, 0.25]$    \\ 
$t$, time                                                                                                   & $\si{T}$                         & \si{s}                      & $[1, 600]$ & $[600, 750]$  \\
\begin{tabular}[c]{@{}c@{}}$T_m$, temperature of medium\end{tabular}                                                                                    & $\Theta$                         & \si{K}                      & $[240, 270]$ & $[270, 280]$    \\  
\begin{tabular}[c]{@{}c@{}}$\Delta_T$, initial sphere temperature \\ 
minus temperature of medium\end{tabular} 
& $\Theta$                         & \si{K}                      & $[50, 80]$ &  $[40, 50]$  \\ 
\begin{tabular}[c]{@{}c@{}}$h_c$, convective heat \\ transfer coefficient\end{tabular}                         & $\si{M} \si{T}^{-3} \Theta^{-1}$ & \si{kg.s^{-3}.K^{-1}}       & \multicolumn{2}{c}{$[100, 160]$} \\ 
$k$, thermal conductivity          & $\si{M} \si{L} \si{T}^{-3} \Theta^{-1}$ & \si{kg.m.s^{-3}.K^{-1}}        & \multicolumn{2}{c}{$[30, 100]$}    \\ 
$c$, specific heat                                                              & $\si{L^2} \si{T^{-2}} \si{\Theta^{-1}}$                  & $\si{m^2.s^{-2}.K^{-1}}$     & \multicolumn{2}{c}{$400$}    \\ 
$\rho$, density  & $\si{M }        \si{L}^{-3}$  & \si{kg.m^{-3}} & \multicolumn{2}{c}{$8000$}   \\ 
\hline
\end{tabular}
\label{tab:solid_sphere_function_settings}
\end{center}
\end{table}

The application is a physical system of the temperature dynamics of a solid sphere at a given distance from its center \citep{cengel2003}. The sphere is immersed in a fluid and has a higher temperature than the fluid when immersed at time 0. The model assumes heat transfer convection between the sphere and the fluid, whereas heat transfer by conduction is assumed within the sphere. The output is temperature of the sphere ($T_s$) in $\si{K}$ (i.e., $[T_s] = \Theta$) which depends on nine inputs; their dimensions, units, and ranges are detailed in Table~\ref{tab:solid_sphere_function_settings}. Note that inputs specific heat ($c$) and density ($\rho$) are kept constant in the system.

\subsection{Dimensional analysis}

\begin{enumerate}[label=\bfseries(\Roman*)] 

\item \textbf{System dimensions.} The inputs and output have $p = 4$ fundamental dimensions: length ($\si{L}$), time ($\si{T}$), temperature ($\Theta$), and mass ($\si{M}$).

\item \textbf{Basis quantities.} 
With $p = 4$ fundamental dimensions, we need to select four basis quantities for FANOVA DA from the set of nine in Table~\ref{tab:solid_sphere_function_settings}. They will lead to $d - p = 5$ derived dimensionless inputs and a dimensionless output.   

We perform FANOVA up to 2-input interactions on 20 repeated experiments (on the seven inputs that vary) with a training size of $n = 70$ runs, the power exponential correlation function, and a constant regression term. Here, the original output $T_s$ and the original inputs in Table~\ref{tab:solid_sphere_function_settings} are used. Then, we obtain the percentage contribution attributed to a main or interaction effect, which is an estimate of its relative importance.

\begin{itemize}

\item \textbf{Main effects.} Figure~\ref{fig:solid_sphere_main_effect_plot} shows boxplots of the percentage contributions to the variance of the predictor for the seven inputs that vary in the system. We can see that, of the four dominant main effects, temperature of medium ($T_m$) and initial sphere temperature minus temperature of medium ($\Delta_T$) involve fundamental dimension $\Theta$ with median percentage contributions of around 30\% and 17.5\%, respectively. The second most important main effect is time ($t$) with a median contribution close to 30\%; it introduces dimension $\si{T}$. Radius of sphere ($r$) has a median contribution of 15\% and involves fundamental dimension $\si{L}$. The final dimension, mass $\si{M}$, appears in $h_c$ and $k$; both have smaller contributions but with $h_c$ consistently slightly larger.

\begin{figure}[ht!]
\begin{center}
\begin{subfigure}{.45\textwidth}
    \centering
    \includegraphics[width=1\textwidth]{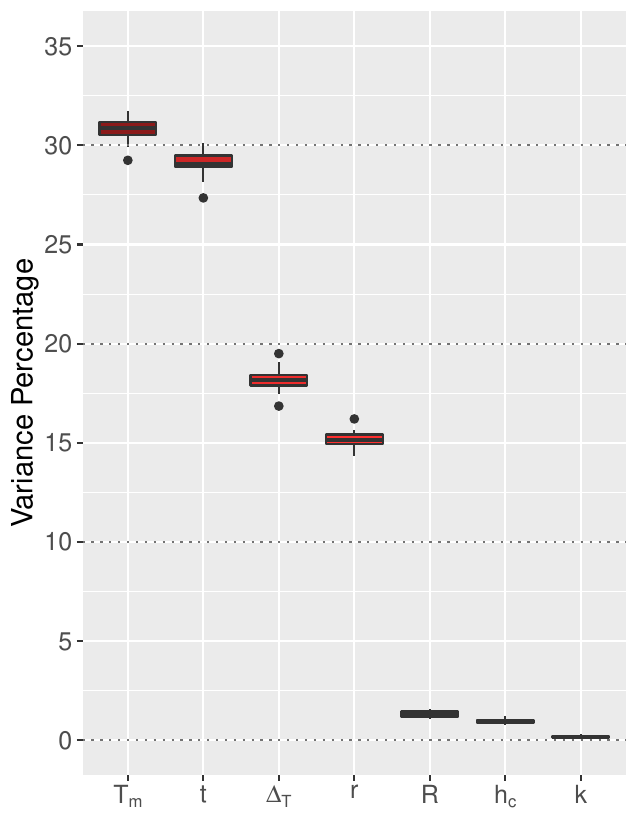}
    \vspace{-0.2\baselineskip}  
    \caption{Main effects.}
    \vspace{1.5\baselineskip}  
    \label{fig:solid_sphere_main_effect_plot}
\end{subfigure}
\begin{subfigure}{.45\textwidth}
     \centering
     \includegraphics[width=1\textwidth]{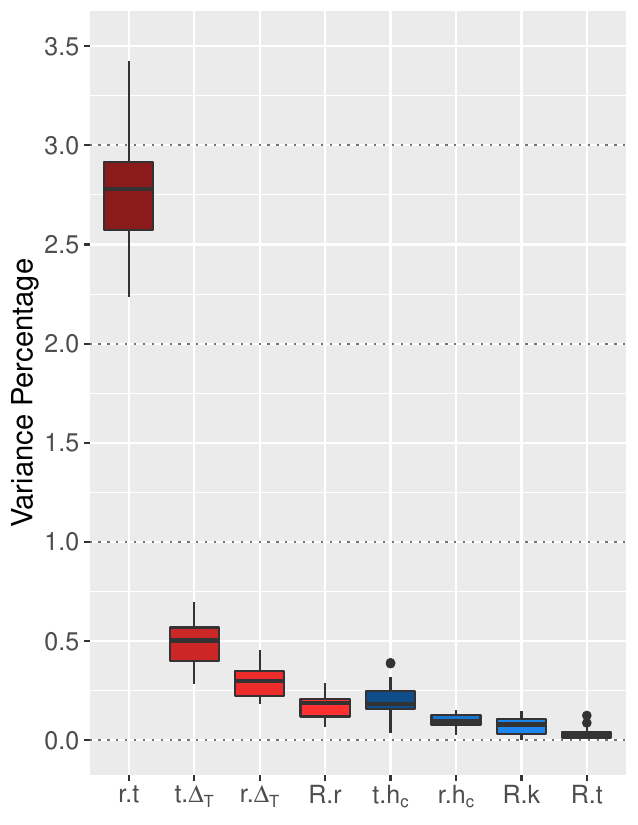}
     \vspace{-0.2\baselineskip}  
     \caption{Input interactions.}
     \vspace{1.5\baselineskip}  
     \label{fig:solid_sphere_Higher_Effects_1_8}
\end{subfigure}
\vspace{-1.3\baselineskip}
\caption{
FANOVA percentage contributions for the solid-sphere function using a GaSP model with power exponential correlation function and a constant regression term. Each boxplot shows results over 20 repeat experiments
with different mLHDs of $n = 70$ runs for training.}
\label{fig:solid_sphere_FANOVA}
\end{center}
\end{figure}

\item \textbf{Input interactions.} The boxplots in Figure~\ref{fig:solid_sphere_Higher_Effects_1_8} of the top eight FANOVA percentages from 2-input interactions do not change the findings from the main-effects plot. The largest contribution, from $r \cdot t$, accounts for a median slightly above 2.75\%, which is consistent with important $r$ and $t$ main effects already identified. The next two largest contributions involve $\Delta_T$, and it could be considered instead of $T_m$ to represent dimension $\Theta$. These interaction contributions are modest, however, and the main effect of $T_m$ substantially dominates $\Delta_T$.

\end{itemize}

Hence, FANOVA suggests the inputs $T_m$, $t$, $r$, and $h_c$ as basis quantities.

\item \textbf{Dimensionless inputs and output.} We will compare two DA approaches: FANOVA DA and the dimensionless quantities used by \cite{tan2017}, abbreviated T-DA.

\begin{itemize}
\item \textbf{T-DA.} 
The quantities used by \cite{tan2017} are the dimensionless output
\begin{equation}
\label{eq:output_DA_Tan}
q_0^{(T)} = \frac{T_s}{\Delta_T},
\end{equation}
and the dimensionless inputs $q_i^{(T)}$ ($i = 1, \ldots, 5$) 
\begin{align}
\label{eq:inputs_DA_Tan}
q_1^{(T)} &= \frac{T_m}{\Delta_T} &  q_2^{(T)} &= R & q_3^{(T)} =& \frac{h_c r}{k} \\
q_4^{(T)} &= \frac{k t}{c \rho r^2} & q_5^{(T)} &= \frac{h_c^2}{\Delta_T c^3 \rho^2} \nonumber.
\end{align}
Here $q_0^{(T)}$ and $q_1^{(T)}$ are dimensionless temperature ratios, $q_2^{(T)}$ is a dimensionless distance ratio from the center of the sphere relative to its radius, $q_3^{(T)}$ is the dimensionless Biot number, $q_4^{(T)}$ is the dimensionless Fourier number, and $q_5^{(T)}$ is the dimensionless convective heat transfer coefficient.

These choices were motivated by the formulation of the system of equations solved in the computer code:
\begin{equation}
\label{eq:equation_sphere}
q_0^{(T)} = q_1^{(T)} + \sum_{i = 1}^{\infty} \frac{4 (\sin \eta_i - \ \eta_i \cos \eta_i)}{2 \eta_i - \sin (2 \eta_i)} e^{-\eta_i^{2} q_4^{(T)}} \frac{\sin \left( \eta_i q_2^{(T)} \right)}{\eta_i q_2^{(T)}};
\end{equation}
where $\eta_i$ is the solution of the equation
\begin{equation*}
1 - \eta_i \cot \eta_i = q_3^{(T)} \quad \text{for } i = 1, \ldots, \infty
\end{equation*}
with $\eta_i \in \left[ (i -1) \pi, i \pi \right]$. Note that, for numerical computations of the output, \cite{tan2017} approximates the series in (\ref{eq:equation_sphere}) with four terms. These equations generate all data used, regardless of the variables appearing in various GaSP models.

\item \textbf{FANOVA DA.} 
There is much latitude here to choose dimensionless quantities, and we are guided by two preliminary analyses. Both take the first training set of $n = 70$ runs for the original variables, without any further data collection motivated by derived quantities.

\begin{figure}[ht!]
\centering
    \includegraphics[width = 0.6\textwidth]{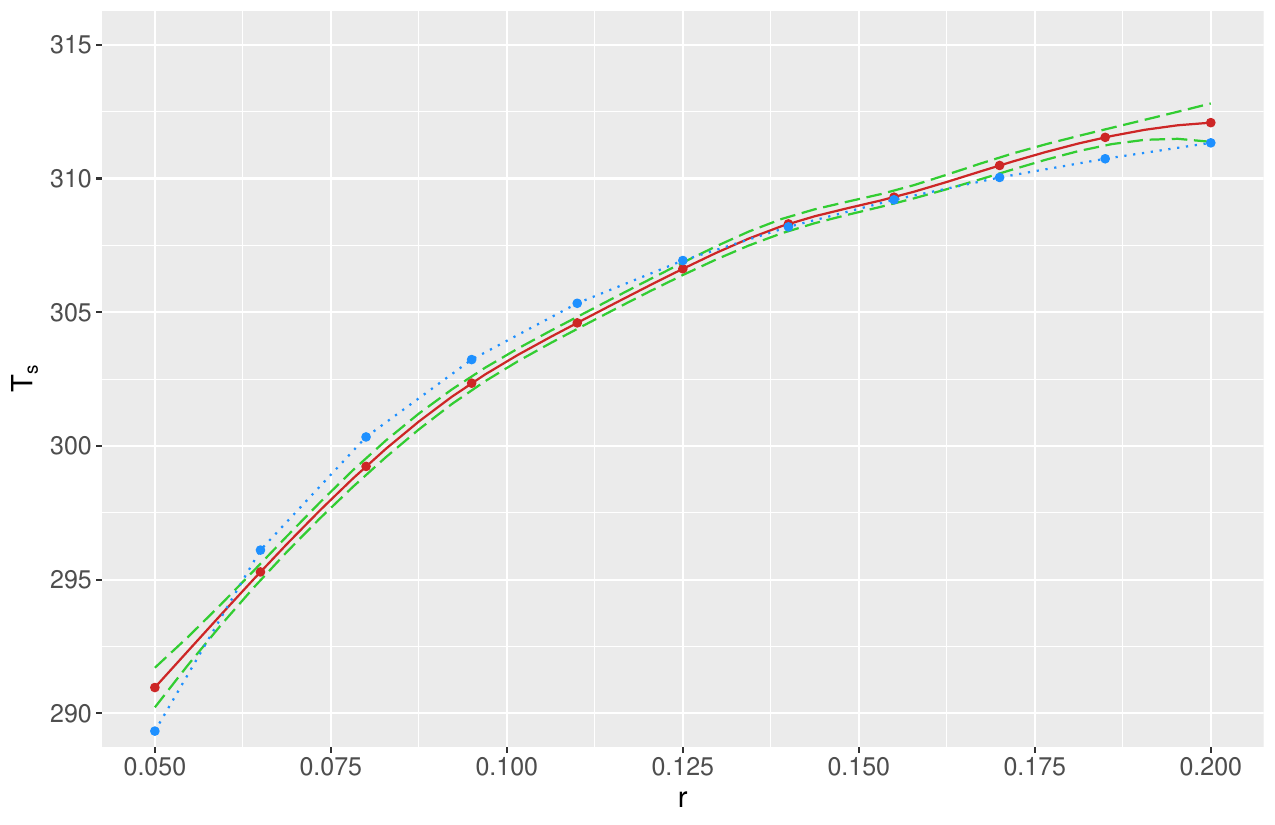}
    \caption{Estimated main effect (solid line) of radius of sphere ($r$) on temperature of sphere ($T_s$) from FANOVA with power exponential correlation function and a constant regression term.
   The dashed lines show approximate pointwise 95\% confidence limits, 
   and the dotted line shows the fitted values from a simple least-squares regression of $T_s$ on $1 / r$.}
\label{fig:solid_sphere_radius}
\end{figure}

First, we can gain insight into the behaviour of the system by visualizing the shapes of the important estimated main effects. As well as computing a scalar percentage FANOVA contribution for a given input, the \texttt{GaSP} package \citep{gasp} can integrate out the remaining variables in the prediction function to visualize the estimated effect of a single input. The estimated main effect of sphere radius $r$ in Figure~\ref{fig:solid_sphere_radius} is the most interesting. It suggests that the effect of increasing $r$ on sphere temperature $T_s$ levels off, which is not surprising as a sphere with a larger radius would cool slower on average. Additionally, the figure includes the fit of a simple linear regression of the estimated main effect on $1/r$. The good fit suggests that $T_s$ is approximately linear in $1/r$ (with a negative slope). Thus, when creating the five dimensionless inputs, $r$ appears in the denominator whenever dimension $L$ is involved.

\begin{figure}[ht!]
\centering
\includegraphics[height=0.7\textheight]{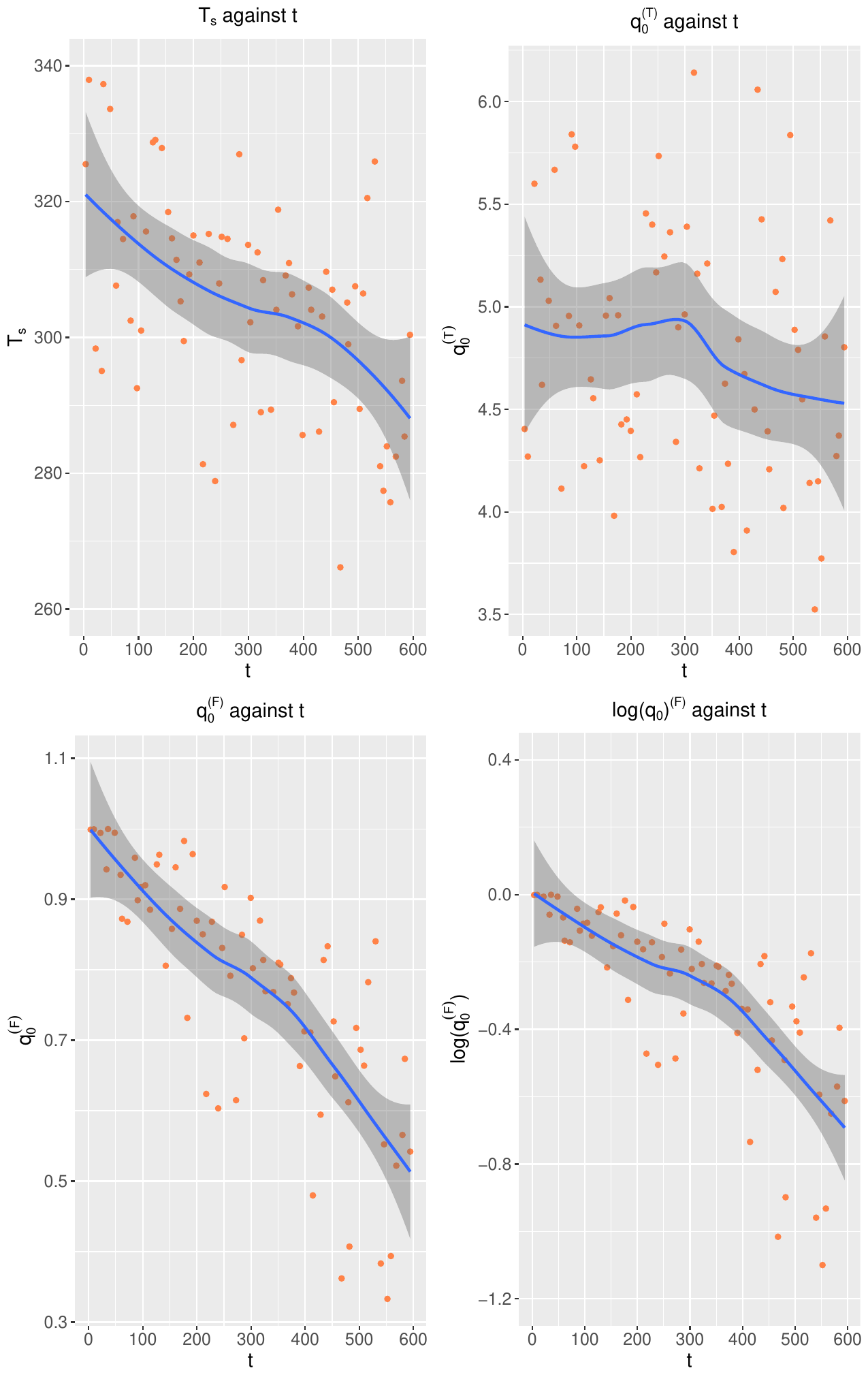}
\caption{Scatterplots of $T_s$ or three derived dimensionless outputs against time ($t$) 
for the solid-sphere function from an mLHD of $n = 70$ runs.}
\label{fig:solid_sphere_scatterplots}
\end{figure}

Secondly, it is also insightful here to consider the dynamic evolution of the output $T_s$ with time $t$.  Figure~\ref{fig:solid_sphere_scatterplots} shows scatter plots of $T_s$, or three dimensionless quantities derived from it, versus $t$, along with the respective locally estimated scatterplot smoothing (LOESS) regressions.

The plot in the first panel of Figure~\ref{fig:solid_sphere_scatterplots} 
of the untransformed (and hence not dimensionless) $T_s$ 
shows a clear downward relationship with $t$, as would be expected from cooling. There is wide scatter in $T_s$ uniformly over the range of input $t$, however, due to the other inputs varying in the design.

Even more scatter and less trend is apparent in the second panel of the first row for the dimensionless output $q_0^{(T)} = T_s / \Delta_T$ in T-DA.  

FANOVA identified $T_m$ as the dominant single input for correcting dimension $\Theta$, and one might consider $T_s / T_m$ for the dimensionless output. While the experiment does not allow $T_m$ or $\Delta_T$ to be close to zero, both $T_s / T_m$ and $T_s / \Delta_T$ could be unstable in these limits, suggesting they are problematic as universal choices. The quantity $q_0^{(F)} = (T_s - T_m) / \Delta_T$ is better behaved for all $\Delta_T > 0$, however: the difference in sphere and medium temperatures must always be less than or equal to the initial difference, and $q_0^{(F)}$ lies on $[0, 1]$. This choice is supported by the important role of $\Delta_T$ identified by FANOVA. The plot in the first panel of the second row of Figure~\ref{fig:solid_sphere_scatterplots}
for $q_0^{(F)}$ shows promise: there is a downward trend with $t$ but with less overall scatter than the panels in the first row, especially for small values of $t$.

Our final choice for the dimensionless output in the GaSP model
reinforces the use of both $T_m$ and $\Delta_T$. \cite{meinsma2019} provided a much simpler example involving the temperature over time of an object in a medium of temperature 0 (presumably $\si{K}$; the author does not give units). According to Newton's law of cooling, the rate of cooling for this object is proportional to the difference between its temperature and the temperature of the medium at time $t$. The solution of the differential equation for this simple and idealized physics suggests $\log \left( q_0^{(F)} \right)$ should be approximately linear in $t$. The final panel in Figure~\ref{fig:solid_sphere_scatterplots} for this derived output is similar to the previous panel as the range of $\log \left( q_0^{(F)} \right)$ is not large, but arguably there is slightly less scatter.
Hence, $\log \left( q_0^{(F)} \right)$ will be the dimensionless output in our DA. Moreover, the LOESS regressions shown in the two bottom panels indicate an approximately linear relationship with $t$, which will be will be important in defining the dimensionless inputs.

Based on these arguments the FANOVA DA inputs are 
\begin{align}
\label{eq:inputs_DA_Full}
q_1^{(F)} &= R &  q_2^{(F)} &= \frac{\Delta_T}{T_m + \Delta_T}  & q_3^{(F)} =& \frac{k}{h_c r} \\
q_4^{(F)} &= \sqrt{\frac{c t^2 T_m}{r^2}} & q_5^{(F)} &= \sqrt[3]{\frac{h_c t^3 T_m}{\rho r^3}}.\nonumber
\end{align}
A few further comments shed light on the details of these choices.

\begin{enumerate}

\item $q_1^{(F)} = R$ is already dimensionless.

\item $q_2^{(F)}$ uses $T_m$ to correct the $\Theta$ dimension of $\Delta_T$. Combining the variables as a proportion of the initial sphere temperature $T_m + \Delta_T$ makes $q_2^{(F)}$ well behaved for all $\Delta_T > 0$.

\item All four dimensions appear in $k$. Dividing by the basis quantities $r$ and $h_c$ removes them, and $r$ in the denominator of $q_3^{(F)}$ is consistent with the conclusion from Figure~\ref{fig:solid_sphere_radius}.

\item The dimensions \si{L}, \si{T}, and \si{\Theta} of $c$ require the three basis quantities $t$, $T_m$, and $r$ for the dimensionless  input $q_4^{(F)}$, and the squared dimensions for length and time in $c$ necessitate $r^2$ and $t^2$. While $c t^2 T_m / r^2$ is already dimensionless, the square root makes $q_4^{(F)}$ linear in $t$ and proportional to $1/r$. Note also that $c$ is constant in the experiment,  but we need $d - p = 5$ derived dimensionless inputs including $q_4^{(F)}$, which is not constant because of the varying basis quantities. 

\item The rationale for  $q_5^{(F)}$ is similar to that for $q_4^{(F)}$. This time a cube root transformation gives $t$ in the numerator and $r$ in the denominator. 

\end{enumerate}

\end{itemize}

\item \textbf{System's transformed function.} 
The transformed functions for the two DA frameworks are therefore
\begin{equation*} 
\frac{T_s - T_m}{\Delta_T} = g \left( R, \frac{\Delta_T}{T_m + \Delta_T}, \frac{k}{h_c r},  
\sqrt{\frac{c t^2 T_m}{r^2}}, \sqrt[3]{\frac{h_c t^3 T_m}{\rho r^3}} \right)
\end{equation*}
for FANOVA DA, and
\begin{equation*}
\frac{T_s}{\Delta_T} = h \left( \frac{T_m}{\Delta_T}, R, \frac{h_c r}{k}, \frac{k t}{c \rho r^2}, 
\frac{h_c^2}{\Delta_T c^3 \rho^2} \right)
\end{equation*}
for T-DA.

\end{enumerate}

\subsection{Simulation settings}

All models are trained with data from 20 repeat experiments for each of $n = 70, 140, 280, 560$. The test set size is fixed at $N = 10,000$,
and again test-set predictions from the DA approaches are transformed back to the original scale when reporting accuracy summaries. In the case of extrapolation, test-set ranges are expanded according to Table~\ref{tab:solid_sphere_function_settings}. The three different approaches, non-DA (the untransformed response and the seven untransformed inputs that vary), T-DA, and FANOVA DA, are all tried with with either constant or linear terms in the regression model.

\subsection{Interpolation}

\begin{center}
\begin{figure}[ht!]
\centering
    \includegraphics[width = 0.97\textwidth]{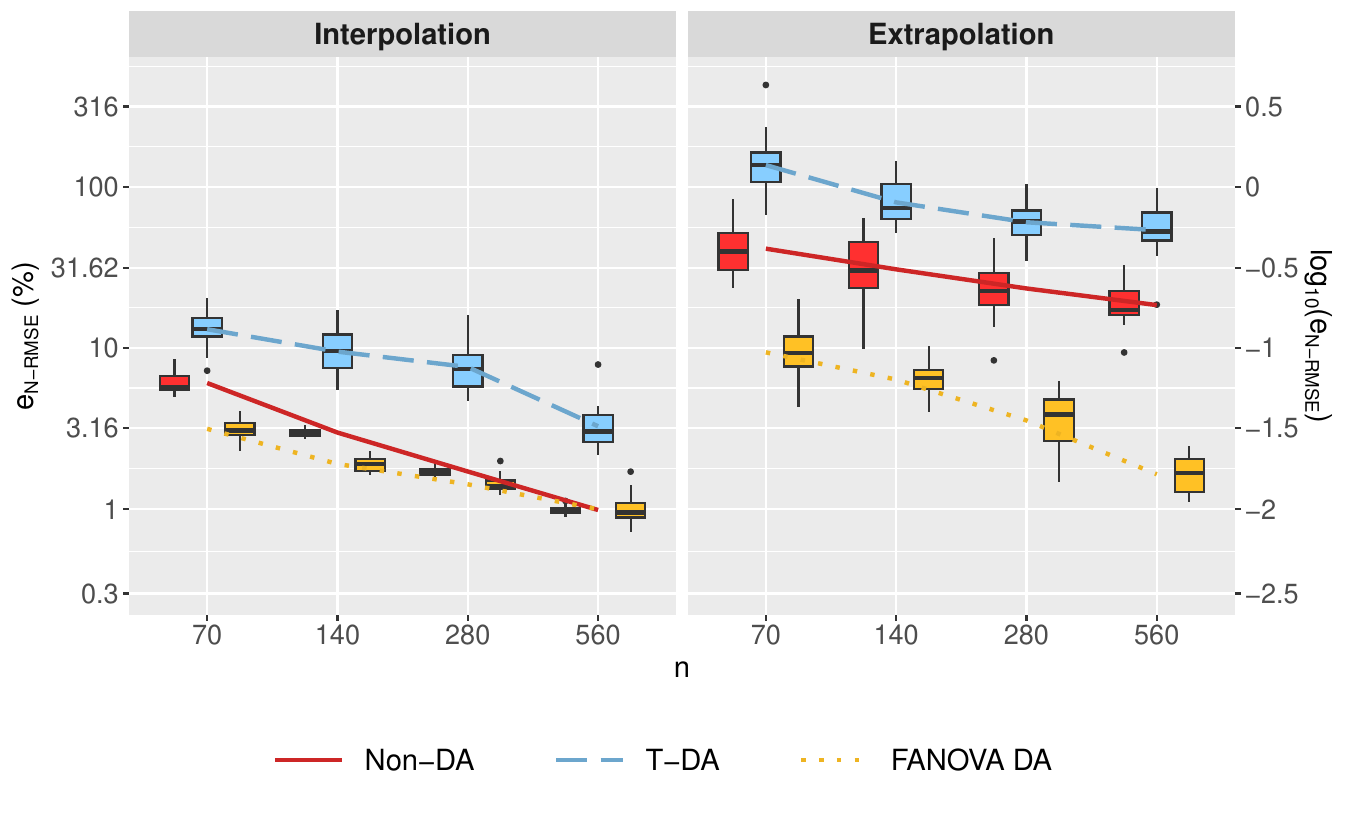}
    \caption{Prediction accuracy by type of DA for the solid-sphere function, with linear regression component. Each boxplot shows results from 20 mLHDs of $n = 70, 140, 280, 560$ runs for training and $N = 10,000$ runs for testing. Input ranges for interpolation and extrapolation in testing are shown in Table~\ref{tab:solid_sphere_function_settings}. The lines join $e_{\mathrm{N-RMSE}}$ sample means.}
\label{fig:sphere:linear}
\end{figure}
\end{center}

Figure~\ref{fig:sphere:linear} compares prediction accuracy for Non-DA, T-DA, and FANOVA DA, where all three methods employ a linear regression component. The results for interpolation in the left panel
show that FANOVA DA provides the best prediction accuracy overall among all the approaches, particularly for small $n$. (Non-DA is competitive at $n = 560$.) T-DA has substantially worse accuracy, even versus non-DA. We note again that T-DA was not developed for GP models, but the findings here emphasize that choice of variables in DA is important.

\subsection{Extrapolation}

The FANOVA results in Figures~\ref{fig:solid_sphere_main_effect_plot} and \ref{fig:solid_sphere_Higher_Effects_1_8} indicate that temperature of medium ($T_m$), time ($t$), initial sphere temperature minus temperature of medium ($\Delta_T$), and radius of sphere ($r$) are the inputs with the most important effects. Hence for challenging extrapolation, their ranges are shifted outside the respective training ranges, as detailed in Table~\ref{tab:solid_sphere_function_settings}.

The right panel of Figure~\ref{fig:sphere:linear} for extrapolation demonstrates that accuracy deteriorates markedly over the new input space compared with interpolation. (Again, all models have a linear regression component.) Nonetheless, there are clear differences between the three approaches. T-DA does not offer competitive accuracy with an average $e_{\mathrm{N-RMSE}}$ of about 56\% at $n = 560$. Non-DA has better but still poor performance with an average $e_{\mathrm{N-RMSE}}$ of about 18\% at the same training size. On the other hand, the prediction performance of FANOVA DA holds up well. Its rate of convergence for $e_{\mathrm{N-RMSE}}$ is about $1 / n^{0.83}$, going from an average $e_{\mathrm{N-RMSE}}$ of around 7\% at $n = 70$ to about 2\% at $n = 560$. Overall, FANOVA DA provides substantially better prediction accuracy for extrapolation.

\subsection{Constant regression}

Prediction accuracy for models with a constant regression component are 
depicted in Figure~\ref{fig:sphere:constant_appendix} of Appendix~\ref{sect:appendix}. Relative to the results in Figure~\ref{fig:sphere:linear}, T-DA improves for interpolation
but is still not competitive with non-DA or FANOVA DA. There is little difference in results for extrapolation due to the different regression models.

For completeness, we also tried an expanded full-log FANOVA DA, similar to the input expansion that worked well for the borehole function. (As throughout for the solid-sphere example,  the  output $\log \left( q_0^{(F)} \right)$ is already on a log scale.) This time we found poor results for interpolated predictions, while FANOVA DA still outperforms the other approaches for extrapolated predictions.

\section{Summary and Discussion}
\label{sect:discussion}

Statistical modeling of a physical system will ideally obey the \cite{buckingham1914} $\Pi$ theorem and hence be a scientifically valid representation. Implementation of the theorem has much latitude, however, and we have demonstrated that user choices matter. DA is not a panacea for a better prediction model; it can actually lead to poorer accuracy. On the other hand, the advantageous choices in this paper gave substantial improvements in accuracy, for interpolation and extrapolation, well beyond the magnitudes of gains often reported for new methods in the analysis of computer experiments.

The implementations of DA in this paper are guided by FANOVA: subject to the constraints of the $\Pi$ theorem, the inputs with the largest contributions to variability in the prediction equation are chosen as basis quantities to remove the dimensions of other inputs and the output. FANOVA's visualization can also guide how these basis quantities are used. Similarly, these ideas could be applied to a physical experiment following a factorial plan, where regular analysis of variance is straightforward.

FANOVA is easy to employ for a computer experiment, as the input domain is usually rectangular in the original inputs. After deriving DA variables, however, the empirical FANOVA approach is difficult to reapply 
using the same data due to the irregularity of the new space. Hence, throughout we used only the original training data for FANOVA and when refitting models in derived variables. Novel designs from a follow-up experiment in a new domain \citep{shen2018_2} may lead to further improvements.

DA requires identification of all variables, even those that are constant like $c$ and $\rho$ in the solid-sphere experiment. A computer experiment has an advantage here, too, as the variables or physical constants in the code are available. Care may be needed, however, with lurking variables, say within description files for a particular setting
(e.g., those defining a fixed topography of an environment).
  
For computer experiments no new software was required for the applications in this article. All model training and FANOVA calculations, including visualization, were performed using the GaSP package \citep{gasp}.
 
Even when the basis quantities are identified, there is still flexibility in how to use them. As a simple example, if $x_1$ is a basis quantity with the same dimensions as $x_2$, should $x_1 / x_2$ or $x_2 / x_1$ be the derived dimensionless input? Both choices would satisfy the $\Pi$ theorem. For the more complex solid-sphere application we again gained insight from FANOVA, using visualization of the input-output relationship to choose functions. For the borehole example, log transformations and input expansion generated many combinations of the basis quantities. In principle that strategy generates more derived variables than necessary according to the $\Pi$ theorem, but it was highly effective.

Finally, we emphasize again the gains seen for extrapolation, a huge challenge for any statistical model. Dimensionless variables are in a sense scale-free and we were able to demonstrate useful prediction accuracy far outside the original training-data ranges.

\section*{Acknowledgements}
We are grateful to Dennis Lin and Chris Nachtsheim for bringing DA to our attention. This research was funded by the Natural Sciences and Engineering Research Council (NSERC) of Canada under Grant RGPIN-2019-05019.

\newpage
\bibliographystyle{apa}
\bibliography{references}

\begin{thebibliography}{}

\bibitem[\protect\astroncite{Albrecht et~al.}{2013}]{AlbNacAlb2013}
Albrecht, M.~C., Nachtsheim, C.~J., Albrecht, T.~A., and Cook, R.~D. (2013).
\newblock Experimental design for engineering dimensional analysis.
\newblock {\em Technometrics}, 55(3):257--270.

\bibitem[\protect\astroncite{Bridgman}{1931}]{bridgman1931}
Bridgman, P.~W. (1931).
\newblock {\em Dimensional Analysis}.
\newblock Yale University Press.

\bibitem[\protect\astroncite{Buckingham}{1914}]{buckingham1914}
Buckingham, E. (1914).
\newblock On physically similar systems; illustrations of the use of
  dimensional equations.
\newblock {\em Phys. Rev.}, 4:345--376.

\bibitem[\protect\astroncite{{\c{C}}engel}{2003}]{cengel2003}
{\c{C}}engel, Y.~A. (2003).
\newblock {\em Heat Transfer: A Practical Approach}.
\newblock McGraw-Hill series in mechanical engineering. McGraw-Hill.

\bibitem[\protect\astroncite{Chen et~al.}{2016}]{chen2016}
Chen, H., Loeppky, J.~L., Sacks, J., and Welch, W.~J. (2016).
\newblock Analysis methods for computer experiments: How to assess and what
  counts?
\newblock {\em Statistical Science}, 31(1):40--60.

\bibitem[\protect\astroncite{Currin et~al.}{1991}]{currin1991}
Currin, C., Mitchell, T., Morris, M., and Ylvisaker, D. (1991).
\newblock Bayesian prediction of deterministic functions, with applications to
  the design and analysis of computer experiments.
\newblock {\em Journal of the American Statistical Association},
  86(416):953--963.

\bibitem[\protect\astroncite{Finney}{1977}]{finney1977}
Finney, D.~J. (1977).
\newblock Dimensions of statistics.
\newblock {\em Journal of the Royal Statistical Society, Series C (Applied
  Statistics)}, 26(3):285--289.

\bibitem[\protect\astroncite{Groemping}{2023}]{DoE.wrapper}
Groemping, U. (2023).
\newblock {\em DoE.wrapper: Wrapper Package for Design of Experiments
  Functionality}.
\newblock R package version 0.12.

\bibitem[\protect\astroncite{Loeppky et~al.}{2009}]{loeppky2009}
Loeppky, J.~L., Sacks, J., and Welch, W.~J. (2009).
\newblock Choosing the sample size of a computer experiment: {A} practical
  guide.
\newblock {\em Technometrics}, 51(4):366--376.

\bibitem[\protect\astroncite{Meinsma}{2019}]{meinsma2019}
Meinsma, G. (2019).
\newblock Dimensional and scaling analysis.
\newblock {\em SIAM Review}, 61(1):159--184.

\bibitem[\protect\astroncite{Morris and Mitchell}{1995}]{morris1995}
Morris, M.~D. and Mitchell, T.~J. (1995).
\newblock Exploratory designs for computational experiments.
\newblock {\em Journal of Statistical Planning and Inference}, 43(3):381--402.

\bibitem[\protect\astroncite{Morris et~al.}{1993}]{morris1993}
Morris, M.~D., Mitchell, T.~J., and Ylvisaker, D. (1993).
\newblock Bayesian design and analysis of computer experiments: {Use} of
  derivatives in surface prediction.
\newblock {\em Technometrics}, 35:243--255.

\bibitem[\protect\astroncite{Sacks et~al.}{1989}]{sacks1989}
Sacks, J., Welch, W.~J., Mitchell, T.~J., and Wynn, H.~P. (1989).
\newblock Design and analysis of computer experiments.
\newblock {\em Statistical Science}, 4(4):409--423.

\bibitem[\protect\astroncite{Schonlau and Welch}{2006}]{schonlau2006}
Schonlau, M. and Welch, W.~J. (2006).
\newblock Screening the input variables to a computer model via analysis of
  variance and visualization.
\newblock In Dean, A. and Lewis, S., editors, {\em Screening: Methods for
  Experimentation in Industry, Drug Discovery, and Genetics}, chapter~14, pages
  308--327. Springer New York.

\bibitem[\protect\astroncite{Shen et~al.}{2014}]{shen2014}
Shen, W., Davis, T., Lin, D., and Nachtsheim, C. (2014).
\newblock Dimensional analysis and its applications in {Statistics}.
\newblock {\em Journal of Quality Technology}, 46(3):185--198.

\bibitem[\protect\astroncite{Shen and Lin}{2018}]{shen2018}
Shen, W. and Lin, D. K.~J. (2018).
\newblock A conjugate model for dimensional analysis.
\newblock {\em Technometrics}, 60(1):79--89.

\bibitem[\protect\astroncite{Shen et~al.}{2018}]{shen2018_2}
Shen, W., Lin, D. K.~J., and Chang, C.-J. (2018).
\newblock Design and analysis of computer experiment via dimensional analysis.
\newblock {\em Quality Engineering}, 30(2):311--328.

\bibitem[\protect\astroncite{Sonin}{2001}]{sonin2001}
Sonin, A.~A. (2001).
\newblock {\em The Physical Basis of Dimensional Analysis}.
\newblock Department of Mechanical Engineering, MIT, Cambridge, MA, USA, 2
  edition.

\bibitem[\protect\astroncite{Stein}{1999}]{stein1999}
Stein, M.~L. (1999).
\newblock {\em Interpolation of Spatial Data: Some Theory for Kriging}.
\newblock Springer Series in Statistics. Springer New York.

\bibitem[\protect\astroncite{Tan}{2017}]{tan2017}
Tan, M. H.~Y. (2017).
\newblock Polynomial metamodeling with dimensional analysis and the effect
  heredity principle.
\newblock {\em Quality Technology \& Quantitative Management}, 14(2):195--213.

\bibitem[\protect\astroncite{Welch and Yang}{2023}]{gasp}
Welch, W.~J. and Yang, Y. (2023).
\newblock {\em GaSP: Train and Apply a Gaussian Stochastic Process Model}.
\newblock R package version 1.0.5.

\bibitem[\protect\astroncite{Worley}{1987}]{worley1987}
Worley, B.~A. (1987).
\newblock Deterministic uncertainty analysis.
\newblock Technical Report CONF-871101-30, Oak Ridge National Lab., TN (USA).

\end{thebibliography}

\newpage
\appendix
\section{Appendix}\label{sect:appendix}

\begin{figure}[ht!]
\begin{center}
\begin{subfigure}{1\textwidth}
\begin{center}
\includegraphics[width=0.9\linewidth]{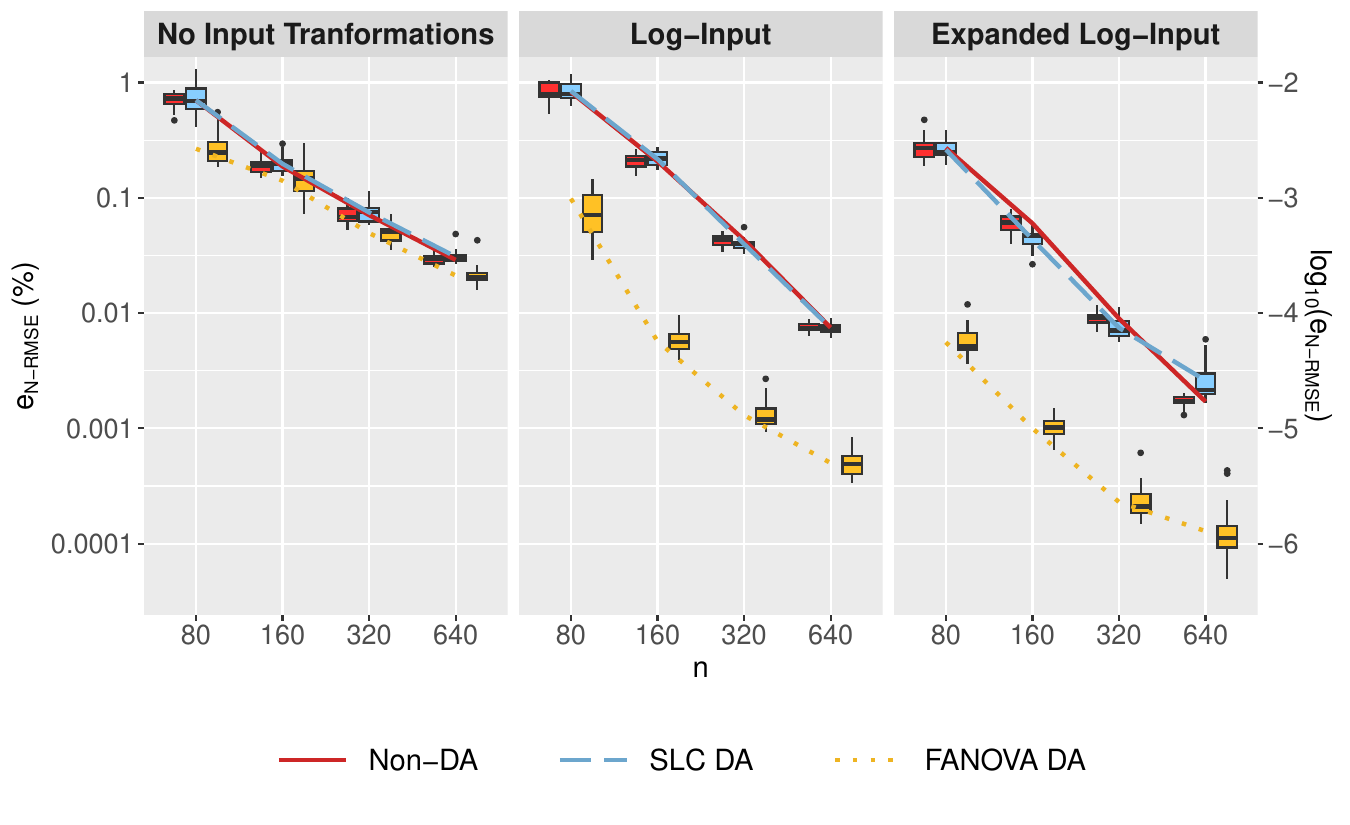}
\vspace{-1\baselineskip}  
\caption{Interpolation.}
\label{fig:b_function_linear_interpolation_plot}
\end{center}
\end{subfigure}
\vspace{1\baselineskip}  
\begin{subfigure}{1\textwidth}
\vspace{1.2\baselineskip} 
\begin{center}
\includegraphics[width=0.9\linewidth]{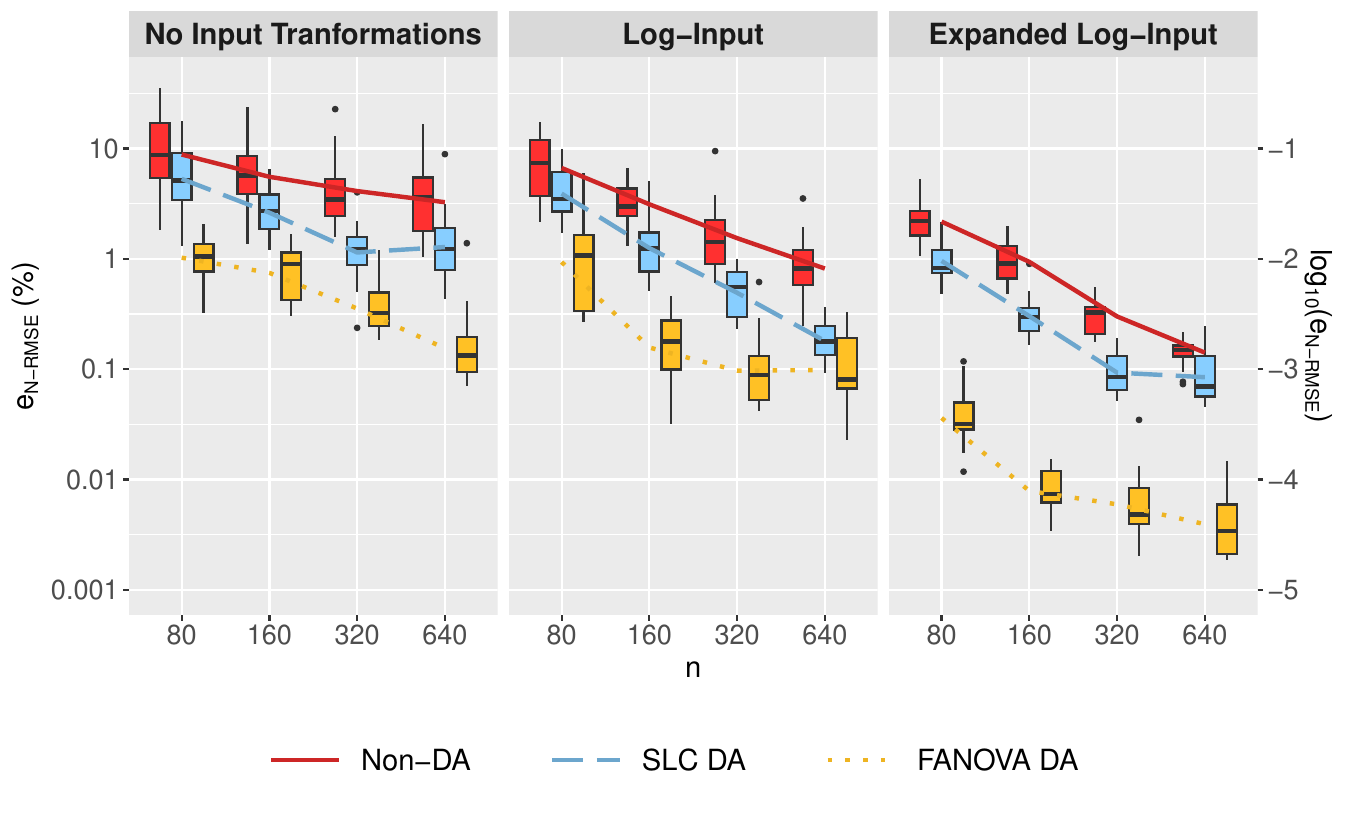}
\vspace{-1\baselineskip} 
\caption{Extrapolation.}
\label{fig:b_function_linear_extrapolation_plot}
\end{center}
\end{subfigure}
\vspace{-0.2\baselineskip}
\caption{Prediction accuracy by type of DA for the borehole function,
    for three input arrangements and a linear regression component. 
   Each boxplot shows results from 20 mLHDs of $n = 80, 160, 320, 640$ runs for training 
   and $N = 10,000$ runs for testing.
   The lines join $e_{\mathrm{N-RMSE}}$ sample means.}
\label{fig:borehole:linear_appendix}
\end{center}
\end{figure}

\newpage
\vspace*{\fill}
\begin{center}
\begin{figure}[ht!]
\centering
    \includegraphics[width = 0.97\textwidth]{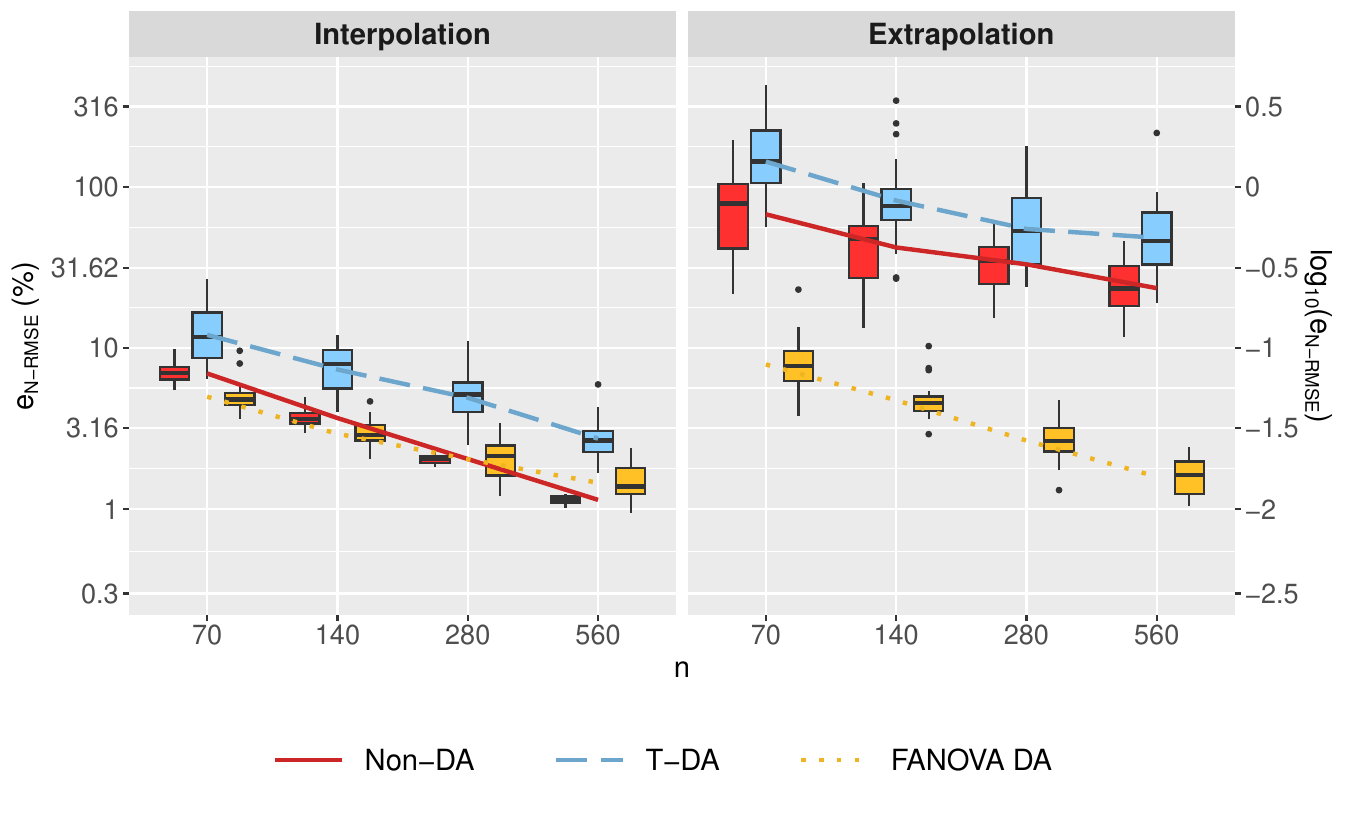}
    \caption{Prediction accuracy by type of DA for the solid-sphere function, with constant regression component. Each boxplot shows results from 20 mLHDs of $n = 70, 140, 280, 560$ runs for training and $N = 10,000$ runs for testing. Input ranges for interpolation and extrapolation in testing are shown in Table~\ref{tab:solid_sphere_function_settings}. The lines join $e_{\mathrm{N-RMSE}}$ sample means.}
\label{fig:sphere:constant_appendix}
\end{figure}
\end{center}
\vspace*{\fill}

\end{document}